\documentclass{article} % For LaTeX2e
\usepackage{main,times}

\usepackage{amsmath,amsfonts,bm}

\def\eqref#1{equation~\ref{#1}}
\def\1{\bm{1}}

\DeclareMathAlphabet{\mathsfit}{\encodingdefault}{\sfdefault}{m}{sl}
\SetMathAlphabet{\mathsfit}{bold}{\encodingdefault}{\sfdefault}{bx}{n}

\usepackage{hyperref}
\usepackage{url}

\usepackage{xspace}
\usepackage{subcaption}
\usepackage{graphicx}
\usepackage{adjustbox}
\usepackage{wrapfig}
\usepackage{tabularx}
\usepackage{multirow}
\usepackage{booktabs}
\usepackage{makecell}

\newcommand{\ours}{ResGVC\xspace}
\newcommand{\ie}{\textit{i.e.}\xspace}

\newcommand{\wrt}{\textit{w.r.t.}\xspace}

\title{ResGVC: Residual Coding for High-fidelity Generative Video Compression}

\author{\textbf{Ding Ding} \quad \textbf{Daowen Li}$\thanks{Project Leader.}$ \quad \textbf{Ying Chen}$\thanks{Corresponding authors.}$ \quad \textbf{Kai Li}$^\dagger$ \\
Alibaba Group \\
    {\tt\small \{liangjie.dd,lidaowen.ldw,YingChen,kaishi.lk\}@alibaba-inc.com}
}

\iclrfinalcopy
\begin{document}

\maketitle

\begin{abstract}
Generative video compression (GVC) methods usually employ decoder-side diffusion models to denoise reconstructed signals based on generative priors. Besides, most of these methods perform denoising in latent space rather than directly in pixel domain.
In this paper, we propose \textbf{\ours}, a residual coding framework for GVC that guides denoising toward original pictures by transmitting residual signals. \ours first encodes the original pictures within a learned video codec, and then decomposes the distortion between the codec reconstruction and the original pictures into a sequence of residual segments, where each residual segment is acquired at each denoising step. Residual segments can further be compressed by a novel residual codec with a diffusion transformer as an entropy model. Unlike prior methods, \ours introduces explicit source constraints throughout denoising rather than relying on decoder-side generation alone. The diffusion-based entropy model also enables more effective distribution modeling capability based on the powerful generative priors.
Moreover, denoising directly in pixel space avoids the irreversible information loss caused by latent representation bottlenecks. 
Experimental results show that, compared with the previous state-of-the-art method, \ours achieves 63.7\% and 83.8\% BD-rate reductions in terms of LPIPS and PSNR, respectively, while also delivering superior visual quality.

\end{abstract}

\section{Introduction}

Traditional video coding standards~\citep{avc,hevc,vvc} and recent neural video compression (NVC) methods~\citep{dcvc,dcvc-fm,dcvc-rt,dcvc-uf} have shown non-trivial improvements in rate-distortion performance. As the optimization prioritizes signal fidelity, these codecs may lose fine textures or produce blurry results at low bitrates, which deviates from human perception. This limitation has motivated generative video compression (GVC), where generative models are integrated into codecs to synthesize plausible details that are not explicitly transmitted.
Recent GVC methods commonly employ decoder-side diffusion models~\citep{sd,cogvideox,wan} to denoise reconstructed signals and improve perceptual quality~\citep{diffvc,s2vc,gnvc-vd,yoda}. Moreover, most work performs denoising in latent space rather than directly in pixel domain.
 
However, in existing GVC methods, the decoder-side denoising process is largely driven by generative priors without sufficient source constraints along the denoising path. As a result, the denoised signal may drift from the original signal during iterative denoising. Besides, because these methods typically perform latent denoising coupled with latent autoencoders, the bottleneck discards fine details that cannot be reliably recovered at the decoder~\citep{vae}. These limitations suggest that faithful GVC requires explicit source constraints throughout denoising and performs denoising directly in pixel space without relying on latent autoencoders.

In this paper, we propose \textbf{\ours}, a \textbf{Res}idual coding framework for high-fidelity \textbf{G}enerative \textbf{V}ideo \textbf{C}ompression, which guides diffusion denoising toward the original pictures by transmitting residual signals. \ours first encodes the original pictures within a learned video codec, and then reformulates the distortion between the codec reconstruction and the original pictures into a sequence of residual segments. Each residual segment is acquired at each denoising step along a predefined denoising path at the encoder side. \ours selectively transmits fidelity-critical residual segments, which serve as source constraints for reconstructing the denoising path at the decoder side. Besides, our denoising is performed directly in pixel space, thereby avoiding the irreversible information loss introduced by latent autoencoders.

\begin{figure*}[t]
\centering
\includegraphics[width=\textwidth]{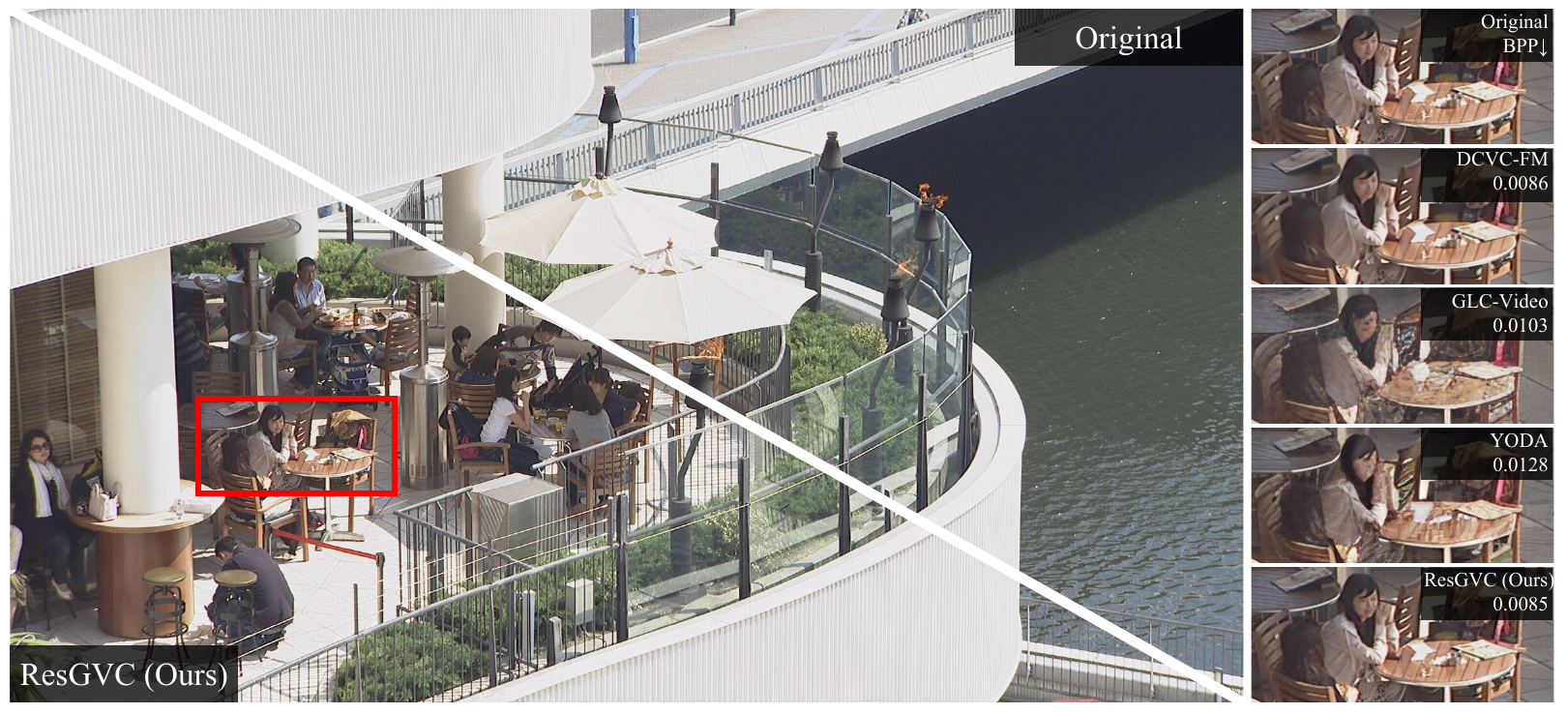}
\vspace{-20pt}
\caption{Visual comparisons on the \textit{BQTerrace} sequence from the HEVC B dataset.}
\vspace{-15pt}
\label{fig:teaser}
\end{figure*}

For efficient residual coding, we propose a residual codec by extending a pixel-space video diffusion transformer from a generator to a conditional probability estimator. This diffusion model estimates the distribution of each residual segment, allowing fidelity-critical residual segments to be entropy-coded while predicting the remaining residual segments from the previously transmitted ones. Thus, residual coding is reserved for information that must be faithfully transmitted to preserve reconstruction quality, whereas residual generation is employed for content that can be well predicted by the generative model to reduce bitrate. By jointly applying residual coding and residual generation along the denoising path, \ours achieves an improved perception-fidelity-rate trade-off.

To the best of our knowledge, this work is the first to explore residual coding in GVC and to repurpose a diffusion model for distribution modeling in both residual coding and residual generation. Experiments demonstrate that \ours achieves state-of-the-art compression performance across multiple metrics. Compared with the latest GVC method, \ours achieves 63.7\% and 83.8\% BD-rate reductions in terms of LPIPS and PSNR, respectively, while producing superior visual results.

Our contributions are summarized as follows:
\begin{itemize}
  \item We propose \ours, a residual coding framework for GVC that guides diffusion denoising toward original pictures by transmitting residuals. It decomposes the distortion between the codec reconstruction and the original pictures into residual segments, which can be transmitted to introduce explicit source constraints across different denoising steps.
  \item We design a novel residual codec with a pixel-space video diffusion transformer as an entropy model. This diffusion model is capable of effective distribution modeling for residual signals, enabling entropy coding for selectively transmitted residual segments while preserving generative capability for non-transmitted ones.
  \item Compared with existing methods, \ours achieves superior compression performance in terms of both perception quality and signal fidelity, while delivering better visual quality.
\end{itemize}

\section{Related work}

\subsection{Generative Video Compression}
Generative video compression (GVC) incorporates generative models into codecs to improve perceptual quality beyond the signal-fidelity objective of traditional video coding and fidelity-oriented NVC. Early GVC methods~\citep{plvc,mentzer2022neural} mainly introduced generative adversarial networks (GANs) and perceptual losses into neural codecs. Although these methods improved perceptual realism, they often suffered from training instability and visual artifacts. Other work explored generative reconstruction on latent or tokenized representations. For example, GLC-Video~\citep{glc} performed compression in latent space of a vector-quantized variational autoencoder, while ProGVC~\citep{li2026progvc} proposed a progressive generative codec based on hierarchical multi-scale residual token maps. However, these methods remained constrained by the representational capacity of their latent or tokenized bottlenecks.

Recently, emerging diffusion models provide strong generative priors for GVC due to their ability to synthesize realistic spatial and temporal details. Existing diffusion-based methods~\citep{diffvc,gnvc-vd,yoda,li2026generative,ma2025diffvc,ma2026diffvc} typically adopt a latent compression pipeline, where the original video is encoded into a
compact latent representation and a diffusion model is employed in the decoder to remove compression distortions from the decoded latent. DiffVC~\citep{diffvc} follows this paradigm by compressing videos in the latent domain and using an image diffusion model to enhance decoded latent frames. GNVC-VD~\citep{gnvc-vd} further introduces a video diffusion transformer to jointly enhance decoded latent representations. YODA~\citep{yoda} explores one-step diffusion-based video compression by embedding multiscale temporal reference features for both latent coding and latent denoising.

Despite their promising perceptual quality, these methods still primarily perform decoder-side denoising in latent space. Their denoising process is largely guided by generative priors, while explicit source constraints along the denoising path remain weak. Moreover, when built upon latent autoencoders, they inherit an irreversible reconstruction bottleneck that discards fine details. In contrast, our method introduces residual coding into pixel-space denoising, enabling explicit source constraints on the denoising process.

\subsection{Diffusion Model with Flow Matching}

Diffusion models have become a popular paradigm for visual generation because they progressively transform noisy samples into clean data using learned denoising dynamics~\citep{ddpm,ddim,sde}. Recent work has developed diffusion architectures based on U-Net~\citep{unet} and transformer~\citep{vaswani2017attention} backbones for high-quality image and video generation~\citep{dit,wan}. To improve efficiency, many methods~\citep{sd,ma2024sit} perform denoising in a compressed latent space which is produced by a variational autoencoder (VAE)~\citep{vae}. While this design substantially reduces computation, it inherits the reconstruction bottleneck of the autoencoder. In contrast, pixel-space diffusion models directly denoise visual signals in the original signal space and can better preserve fine-grained details~\citep{jit,pixnerd,pixeldit,ma2026pixelgen}.

Flow matching~\citep{lipman2023flowmatching,liu2022rectifiedflow} provides an effective formulation for diffusion-based generative modeling by learning a continuous denoising path from a random noise sample to clean data. Specifically, the intermediate noisy state at denoising step $t$ is commonly defined as
\begin{equation}
\mathbf{x}_t=(1-\sigma_t)\mathbf{x}_0+\sigma_t\mathbf{x}_T,\quad 0\le t\le T,\label{eq:forward process}
\end{equation}
where $\mathbf{x}_0$ denotes the clean data. $\mathbf{x}_T$ denotes the denoising start, which is usually a random noise sample. $\sigma_t$ controls the noise level~\citep{edm}. The denoising network is trained to predict the corresponding velocity field by minimizing
\begin{equation}
\mathcal{L}_{\mathrm{FM}}=\mathbb{E}_{t,\mathbf{x}_0,\mathbf{x}_T}\left[
\left\|\hat{\boldsymbol{\epsilon}}(\mathbf{x}_t,\mathbf{c},t)-(\mathbf{x}_T-\mathbf{x}_0)\right\|_2^2
\right], \label{eq:flow matching loss}
\end{equation}
where $\mathbf{c}$ denotes the condition and $\hat{\boldsymbol{\epsilon}}(\mathbf{x}_t,\mathbf{c},t)$ is the predicted velocity. During inference, the denoising network iteratively updates the intermediate noisy state along the predicted denoising path toward the clean data:
\begin{equation}
\hat{\mathbf{x}}_{t-1}=\hat{\mathbf{x}}_t+(\sigma_{t-1}-\sigma_t)\hat{\boldsymbol{\epsilon}}(\hat{\mathbf{x}}_t,\mathbf{c},t),\label{eq:denoising network prediction}
\end{equation}
where $\hat{\mathbf{x}}_{t}$ represents the predicted noisy state at denoising step $t$.

In GVC, diffusion models using flow matching are mainly employed as decoder-side denoisers to refine reconstructed latent representations with generative priors~\citep{diffvc,gnvc-vd,yoda,li2026generative,ma2025diffvc}. Under this formulation, restoration can be considered as estimating a denoising path from the codec reconstruction toward the original pictures. Existing methods primarily predict the denoising path from generative priors, without sufficient source constraints. This motivates our design: instead of treating the denoising path as purely generative, we use transmitted residual signals to guide the denoising process, thereby constraining it with explicit source information. Moreover, unlike latent-space diffusion used in most prior methods, our approach performs denoising in pixel space without the latent representation bottleneck.

\section{Methodology}

\ours represents a residual coding framework for high-fidelity GVC. It first compresses original pictures within a learned video codec~\citep{dcvc-uf}, and then refines the decoded pictures using residual information through iterative residual coding. The whole diffusion process is performed directly in pixel domain, avoiding the irreversible bottleneck of latent space. Concretely, the distortion between the decoded pictures and the original pictures is decomposed into a sequence of residual segments, each obtained at a denoising step. To encode these segments efficiently, we design a residual codec in which a pixel-space video diffusion model serves as a conditional probability estimator for entropy coding. Section~\ref{subsec:bridging} formulates the residual segments from the codec reconstruction to the original pictures. Section~\ref{subsec:framework} presents the proposed residual codec and its diffusion-based entropy model. Section~\ref{subsec:switching} describes a skip mode for residual coding, which skips residual coding at selected denoising steps and replaces it with decoder-side residual generation for reducing bitrate.

\subsection{From Denoising Path to Residual Coding}\label{subsec:bridging}

\begin{figure*}[t]
\centering
\includegraphics[width=\textwidth]{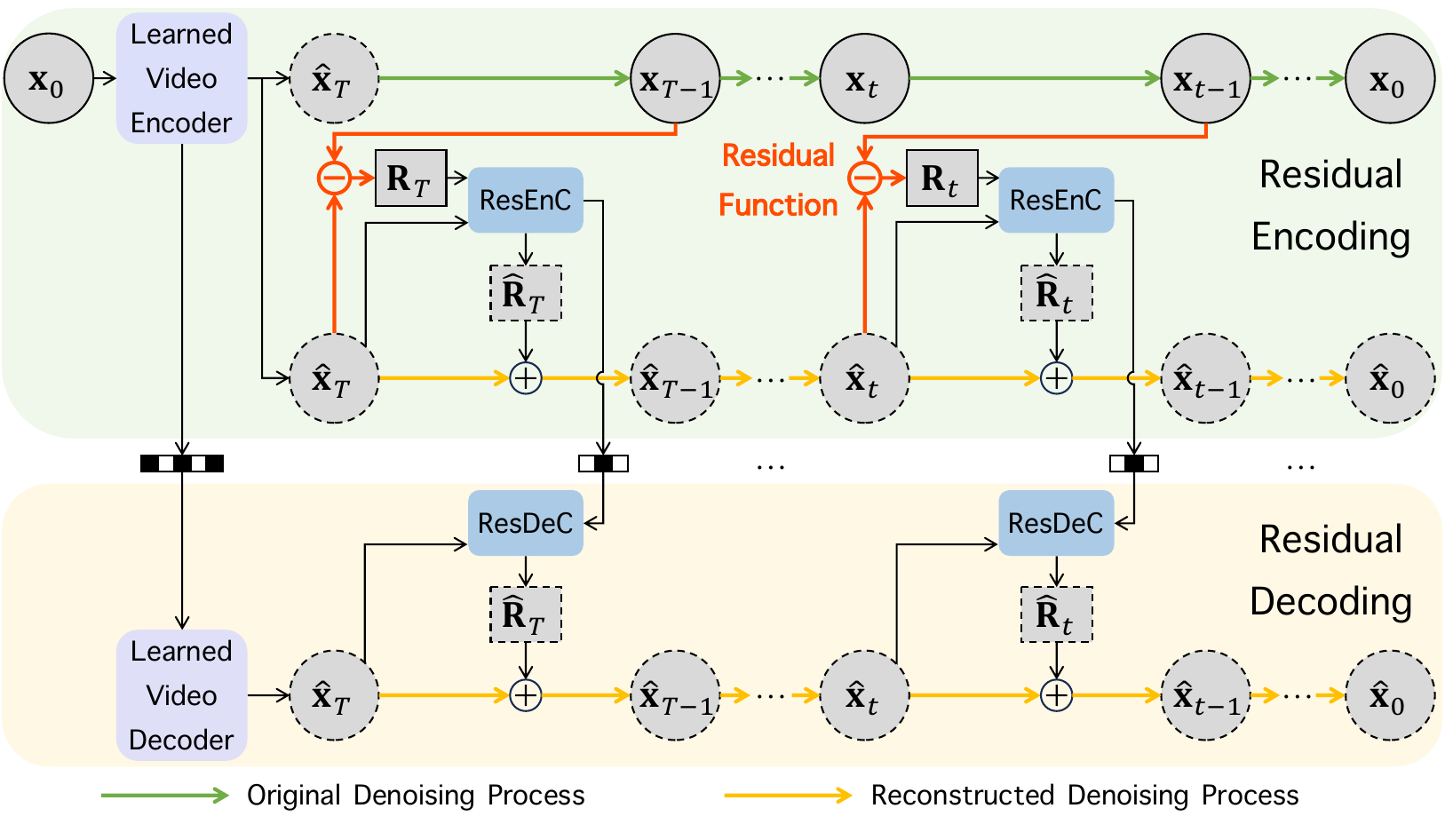}
\vspace{-20pt}
\caption{Overview of the proposed residual coding framework. $\mathbf{x}_0$ denotes the original pictures, and $\hat{\mathbf{x}}_{T}$ denotes the reconstruction decoded by the learned video codec. $\mathbf{x}_{t}$ and $\hat{\mathbf{x}}_t$ denote the predefined and reconstructed noisy states at denoising step $t$, respectively. $\mathbf{R}_t$ denotes the to-be-encoded residual segment, and $\hat{\mathbf{R}}_t$ denotes its reconstruction produced by the residual codec. $\hat{\mathbf{x}}_0$ denotes the final reconstructed pictures after residual decoding. ``ResEnC'' and ``ResDeC'' are the proposed residual encoder and decoder, respectively.}
\vspace{-10pt}
\label{fig:framework}
\end{figure*}

Fig.~\ref{fig:framework} illustrates the proposed framework for residual coding. Given the original pictures $\mathbf{x}_0$, $\hat{\mathbf{x}}_{T}$ denotes the reconstruction decoded by the learned video codec. In diffusion-based GVC, denoising usually starts from $\hat{\mathbf{x}}_{T}$ rather than pure Gaussian noise~\citep{gnvc-vd}. Following flow matching, the predefined noisy state at step $t$ in Eq.~\ref{eq:forward process} is rewritten as
\begin{equation}
\mathbf{x}_t=(1-\sigma_t)\mathbf{x}_0+\sigma_t\hat{\mathbf{x}}_{T},
\label{eq:predefined_state}
\end{equation}
where $\sigma_t$ denotes the noise level of the intermediate noisy state at step $t$. The sequence $\{\mathbf{x}_t\}_{t=0}^{T-1}$ defines the \emph{original denoising process} from the codec reconstruction $\hat{\mathbf{x}}_{T}$ to the original data $\mathbf{x}_0$. Accordingly, the \emph{original denoising path} is formulated as $\hat{\mathbf{x}}_{T}\rightarrow\mathbf{x}_{T-1}\rightarrow\cdots\rightarrow\mathbf{x}_{t}\rightarrow\cdots\rightarrow\mathbf{x}_{0}$.

Prior diffusion-based methods recover this denoising path at the decoder by predicting the denoising update at each step. From step $t$ to step $t-1$, the update in Eq.~\ref{eq:denoising network prediction} can be rewritten as
\begin{equation}
\hat{\mathbf{x}}_{t-1}
=
\hat{\mathbf{x}}_{t}
+
(\sigma_{t-1}-\sigma_t)\hat{\mathbf{R}}_{t},
\label{eq:denoising_update}
\end{equation}
where $\hat{\mathbf{x}}_t$ and $\hat{\mathbf{R}}_{t}$ denote the reconstructed noisy state and the predicted residual information at step $t$, respectively. $(\sigma_{t-1}-\sigma_t)\hat{\mathbf{R}}_{t}$ represents the correction applied to the current reconstructed noisy state to approach the next cleaner state. Since this correction is predicted solely from generative priors, the reconstructed denoising process may deviate from the original denoising process.

In contrast, as illustrated in Fig.~\ref{fig:framework}, we reformulate the same stepwise denoising update into residual coding, which aims to faithfully
approximate the original denoising path at the decoder based on the transmitted residual information. At step $t$, the to-be-encoded residual segment is defined as the scaled difference between the predefined noisy state $\mathbf{x}_{t-1}$ and the reconstructed noisy state $\hat{\mathbf{x}}_t$:
\begin{equation}
\mathbf{R}_t=(\mathbf{x}_{t-1}-\hat{\mathbf{x}}_t) / (\sigma_{t-1}-\sigma_t),
\label{eq:coding_target}
\end{equation}
where the scaling factor $(\sigma_{t-1} - \sigma_t)$ standardizes the variance of $\mathbf{R}_t^i$, thereby aligning its distributions across steps to facilitate robust probability estimation in the entropy model. $(\sigma_{t-1}-\sigma_t)\mathbf{R}_t$ represents the correction that moves $\hat{\mathbf{x}}_t$ toward $\mathbf{x}_{t-1}$, \ie, the next predefined noisy state on the original denoising path. The sequence $\{\hat{\mathbf{x}}_t\}_{t=0}^{T-1}$ defines the \emph{reconstructed denoising process}. Accordingly, the \emph{reconstructed denoising path} is formulated as $\hat{\mathbf{x}}_{T}\rightarrow\hat{\mathbf{x}}_{T-1}\rightarrow\cdots\rightarrow\hat{\mathbf{x}}_{t}\rightarrow\cdots\rightarrow\hat{\mathbf{x}}_{0}$.

At the encoder, each $\mathbf{R}_t$ is compressed into a bitstream, and then reconstructed as $\hat{\mathbf{R}}_t$, which is used to simulate the decoder-side reconstructed denoising process. Starting from $\hat{\mathbf{x}}_{T}$, the encoder iteratively updates $\hat{\mathbf{x}}_t$ by applying Eq.~\ref{eq:denoising_update}, which yields the reconstructed denoising process $\{\hat{\mathbf{x}}_t\}_{t=0}^{T-1}$. 

At the decoder, the reconstructed denoising process also starts from the same codec reconstruction $\hat{\mathbf{x}}_{T}$. The reconstructed residual segments $\{\hat{\mathbf{R}}_t\}_{t=1}^{T}$ are sequentially decoded from the transmitted bitstream, for which the reconstructed denoising process can be reproduced step by step, according to Eq.~\ref{eq:denoising_update}. In this way, our method turns denoising from a purely decoder-side generation process into a high-fidelity reconstruction process constrained by transmitted residual information.

\subsection{Residual Coding with Diffusion-based Entropy Model}\label{subsec:framework}

\begin{figure}[t]
\centering
\begin{subfigure}{0.325\linewidth}
    \centering
    \includegraphics[width=\linewidth]{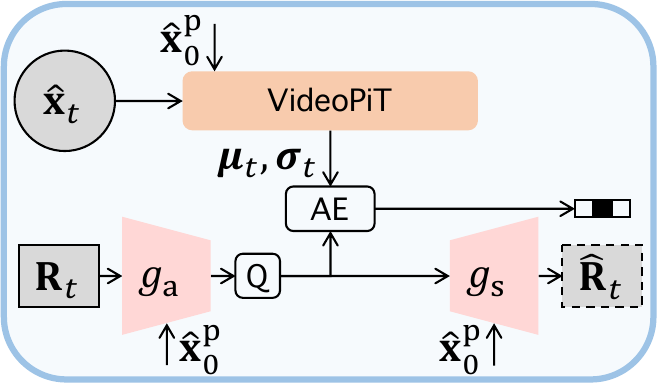}
    \caption{Residual Encoder (ResEnC).}
    \label{subfig:residual encoder}
\end{subfigure}
\hfill
\begin{subfigure}{0.325\linewidth}
    \centering
    \includegraphics[width=\linewidth]{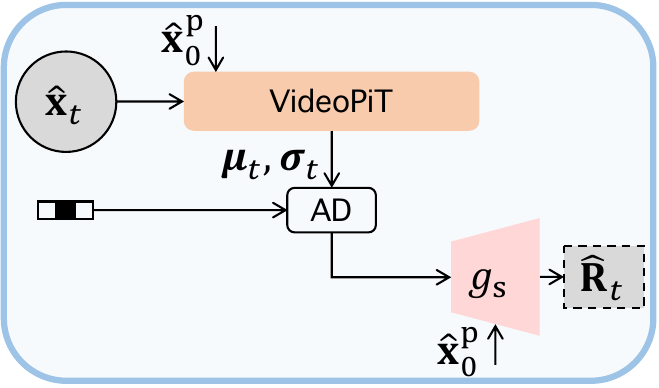}
    \caption{Residual Decoder (ResDeC).}
    \label{subfig:residual decoder}
\end{subfigure}
\hfill
\begin{subfigure}{0.325\linewidth}
    \centering
    \includegraphics[width=\linewidth]{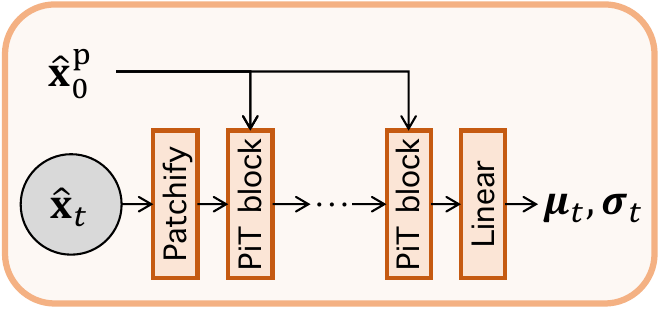}
    \caption{Video Pixel Diffusion Transformer (VideoPiT).}
    \label{subfig:VideoPiT}
\end{subfigure}
\vspace{-5pt}
\caption{Proposed residual encoder, residual decoder, and video pixel diffusion transformer.}
\label{fig:residual encoder and decoder}
\vspace{-10pt}
\end{figure}

Fig.~\ref{fig:residual encoder and decoder} shows the proposed residual codec which consists of a residual encoder (ResEnC) and a residual decoder (ResDeC), with a shared diffusion-based entropy model implemented by a video pixel diffusion transformer (VideoPiT). Generally, $\mathbf{x}_{0}$ can represent either a group of pictures (GOP)~\citep{dcvc-uf} or a single picture~\citep{dcvc-rt}. We define $\mathbf{x}_{0}$ as a GOP and adopt DCVC-UF~\citep{dcvc-uf} as the learned video codec to encode and decode it, yielding $\hat{\mathbf{x}}_{T}$. The residual codec progressively transmits the residual segments $\{\mathbf{R}_t\}_{t=1}^{T}$ between  $\hat{\mathbf{x}}_{T}$ and $\mathbf{x}_{0}$, for which the reconstructed denoising process is recovered step by step, ultimately yielding the final reconstruction $\hat{\mathbf{x}}_0$. 

At the encoder, each residual segment $\mathbf{R}_t$ is first transformed as below:
\begin{equation}
\mathbf{y}_t=g_\mathrm{a}(\mathbf{R}_t\mid \hat{\mathbf{x}}_0^{\mathrm{p}}),
\label{eq:analysis_transform}
\end{equation}
where $g_\mathrm{a}$ denotes the analysis transform built upon the encoder architecture from~\citet{dcvc-uf}. Implementation details are provided in the appendix. $\hat{\mathbf{x}}_0^{\mathrm{p}}$ is the final reconstruction of the previous GOP which is used as temporal context. The transformed representation $\mathbf{y}_t$ is then quantized as
\begin{equation}
\hat{\mathbf{y}}_t=\lfloor \mathbf{y}_t\rceil.
\label{eq:quantization}
\end{equation}
To entropy-code $\hat{\mathbf{y}}_t$, we estimate its conditional probability distribution from the reconstructed noisy state $\hat{\mathbf{x}}_t$ in the reconstructed denoising process. Specifically, $\hat{\mathbf{x}}_t$ is fed into the diffusion-based entropy model, conditioned on the final reconstruction of the previous GOP $\hat{\mathbf{x}}_0^{\mathrm{p}}$, which predicts the mean and scale parameters (\ie, $\boldsymbol{\mu}_t,\boldsymbol{\sigma}_t$) of a Gaussian probability distribution:
\begin{equation}
\boldsymbol{\mu}_t,\boldsymbol{\sigma}_t
=
\operatorname{VideoPiT}(\hat{\mathbf{x}}_t\mid \hat{\mathbf{x}}_0^{\mathrm{p}}),
\label{eq:probability_estimation}
\end{equation}
where $\hat{\mathbf{x}}_t$ is first patchified into tokens, and subsequently processed by a cascade of PiT blocks~\citep{pixeldit} with $\hat{\mathbf{x}}_0^{\mathrm{p}}$ as the condition. The resulting features are projected by a linear layer to generate $\boldsymbol{\mu}_t$ and $\boldsymbol{\sigma}_t$ for entropy coding. More implementation details of the VideoPiT are provided in the appendix. $\hat{\mathbf{y}}_t$ is then arithmetic-encoded as
\begin{equation}
B_t=\operatorname{AE}(\hat{\mathbf{y}}_t;\boldsymbol{\mu}_t,\boldsymbol{\sigma}_t),
\label{eq:arithmetic_encoding}
\end{equation}
where $B_t$ denotes the bitstream of the residual segment at step $t$. $\operatorname{AE}$ denotes the arithmetic encoding. During training, the quantization and arithmetic coding operations are replaced by differentiable approximations for end-to-end optimization, following the GVC practice~\citep{gnvc-vd}.

At the decoder, each residual segment is reconstructed from the transmitted bitstream using the same conditional probability model. Since the decoder has access to the same conditioning information $\hat{\mathbf{x}}_t$ and $\hat{\mathbf{x}}_0^{\mathrm{p}}$ as the encoder, the Gaussian probability parameters in Eq.~\ref{eq:probability_estimation} can be reproduced exactly at the decoder side. The received bitstream is first arithmetic-decoded as
\begin{equation}
\hat{\mathbf{y}}_t
=
\operatorname{AD}(B_t;\boldsymbol{\mu}_t,\boldsymbol{\sigma}_t),
\label{eq:arithmetic_decoding}
\end{equation}
where $\operatorname{AD}$ denotes the arithmetic decoding. $\hat{\mathbf{y}}_t$ is then mapped back to $\hat{\mathbf{R}}_t$:
\begin{equation}
\hat{\mathbf{R}}_t
=
g_\mathrm{s}(\hat{\mathbf{y}}_t\mid \hat{\mathbf{x}}_0^{\mathrm{p}}),
\label{eq:synthesis_transform}
\end{equation}
where $g_\mathrm{s}$ denotes the synthesis transform. Once $\hat{\mathbf{R}}_t$ is reconstructed, it is used to update the current reconstructed noisy state $\hat{\mathbf{x}}_{t}$ toward the cleaner one $\hat{\mathbf{x}}_{t-1}$ using Eq.~\ref{eq:denoising_update}. Note that $g_\mathrm{s}$ is employed in both ResEnC and ResDeC to obtain the same $\hat{\mathbf{R}}_t$, as shown in Fig.~\ref{subfig:residual encoder} and Fig.~\ref{subfig:residual decoder}. Therefore, $\hat{\mathbf{R}}_t$ can be used at the encoder side to construct the next residual segment according to  Eq.~\ref{eq:coding_target}, while simulating the decoder-side reconstructed denoising process. Starting from $\hat{\mathbf{x}}_{T}$, the decoder sequentially recovers all residual segments and reproduces the reconstructed denoising process to obtain $\hat{\mathbf{x}}_0$. 

\subsection{Skip mode for residual coding}\label{subsec:switching}

\begin{figure}[t]
\centering
\begin{subfigure}{0.485\linewidth}
    \centering
    \includegraphics[width=\linewidth]{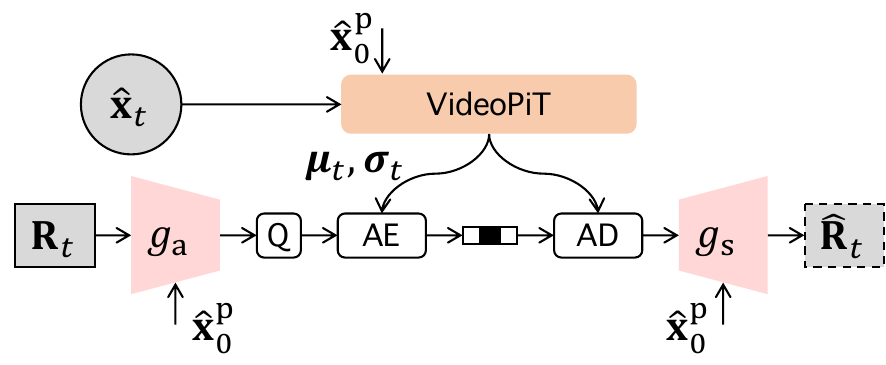}
    \caption{Residual Coding.}
    \label{fig:residual codec}
\end{subfigure}
\hfill
\begin{subfigure}{0.485\linewidth}
    \centering
    \includegraphics[width=\linewidth]{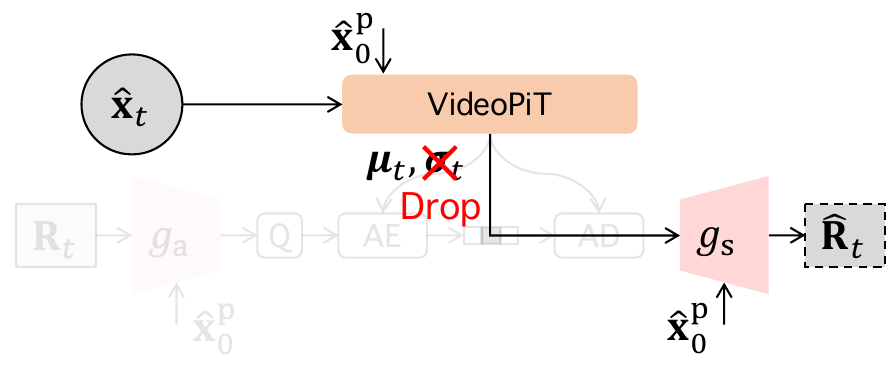}
    \caption{Residual Generation.}
    \label{fig:residual generator}
\end{subfigure}
\vspace{-5pt}
\caption{Residual coding and residual generation under the skip mode.}
\label{fig:residual coding and generation}
\vspace{-10pt}
\end{figure}

\ours supports a skip mode, in which residual coding is applied only to selected denoising steps. For the skipped steps, the corresponding residual segments are not transmitted but are instead generated at the decoder without incurring additional bitrate or quantization error, as illustrated in Fig.~\ref{fig:residual coding and generation}. Specifically, each residual segment  $\hat{\mathbf{R}}_t$ can be recovered at the decoder in one of two modes: \emph{residual coding}, in which $\hat{\mathbf{R}}_t$ is reconstructed from the bitstream, and \emph{residual generation}, in which $\hat{\mathbf{R}}_t$ is predicted by the diffusion model, \ie, VideoPiT. Residual coding is employed in the earlier denoising steps, which are more crucial to the final reconstruction quality, to faithfully transmit fidelity-critical residual segments. In the later steps, by leveraging the informative context from previously decoded residual signals, VideoPiT serves as a generator to estimate $\boldsymbol{\mu}_t$ and $\boldsymbol{\sigma}_t$ without bitstream constraints. The estimated $\boldsymbol{\mu}_t$ is directly used as the predicted $\hat{\mathbf{R}}_t$, while $\boldsymbol{\sigma}_t$ is discarded. Therefore, given a total of $T$ denoising steps, the number of residual-coding steps determines the trade-off between bitrate and reconstruction quality, and its effect is further analyzed in Section~\ref{subsec:ablation}.

Although transmitted and generated residual segments are obtained in different ways, they are incorporated using the same state update rule in Eq.~\ref{eq:denoising_update}. Therefore, transmitted source information and generative priors jointly guide the reconstructed denoising path toward the original pictures.

\section{Experiment}
\begin{figure*}[t]
  \centering

  \begin{subfigure}{0.25\linewidth}
    \includegraphics[width=\textwidth]{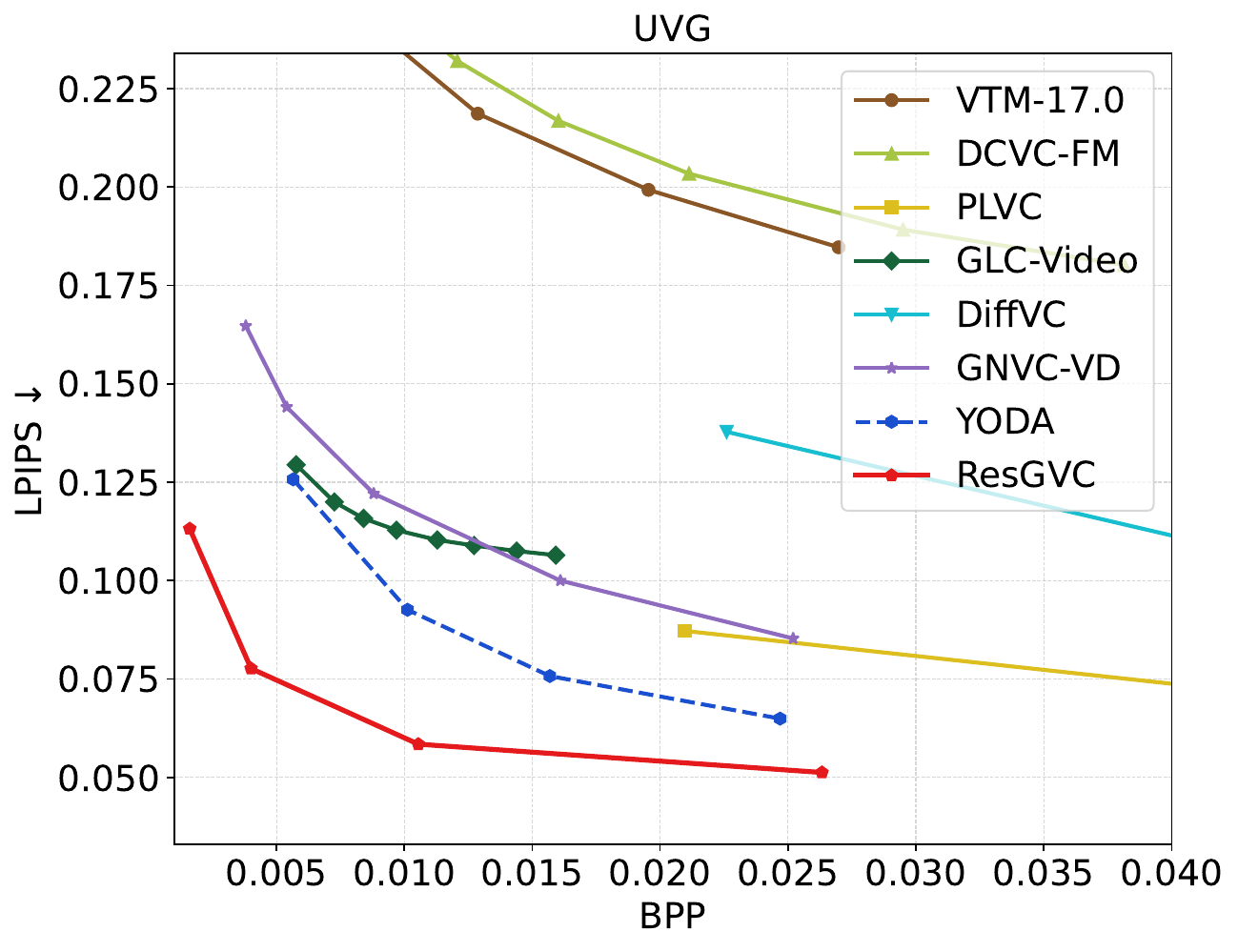}
    \label{fig:lpips_UVG}
  \end{subfigure}
    \hfill
    \hspace{-10pt}
  \begin{subfigure}{0.25\linewidth}
    \includegraphics[width=\textwidth]{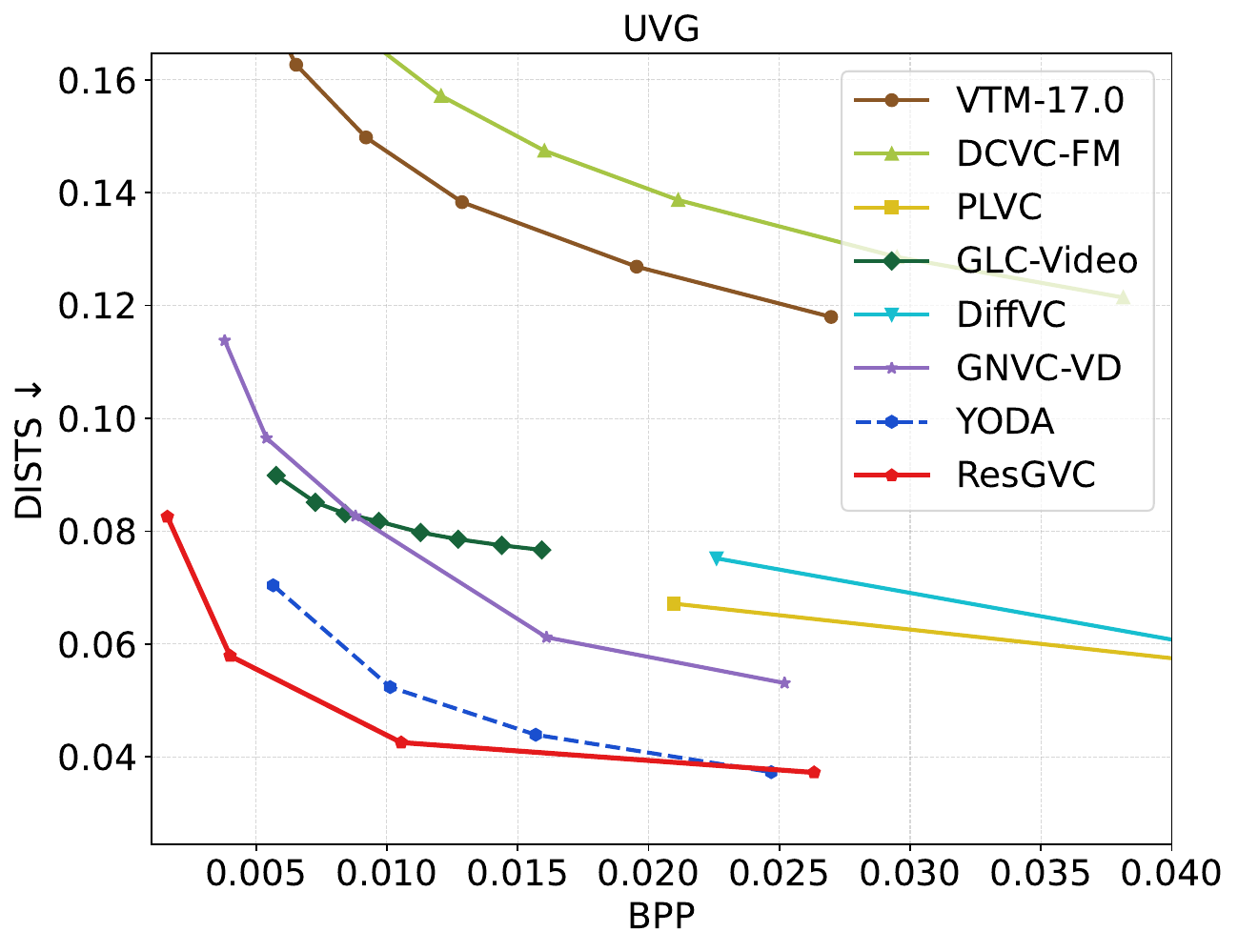}
    \label{fig:dist_UVG}
  \end{subfigure}
      \hfill
      \hspace{-10pt}
  \begin{subfigure}{0.25\linewidth}
    \includegraphics[width=\textwidth]{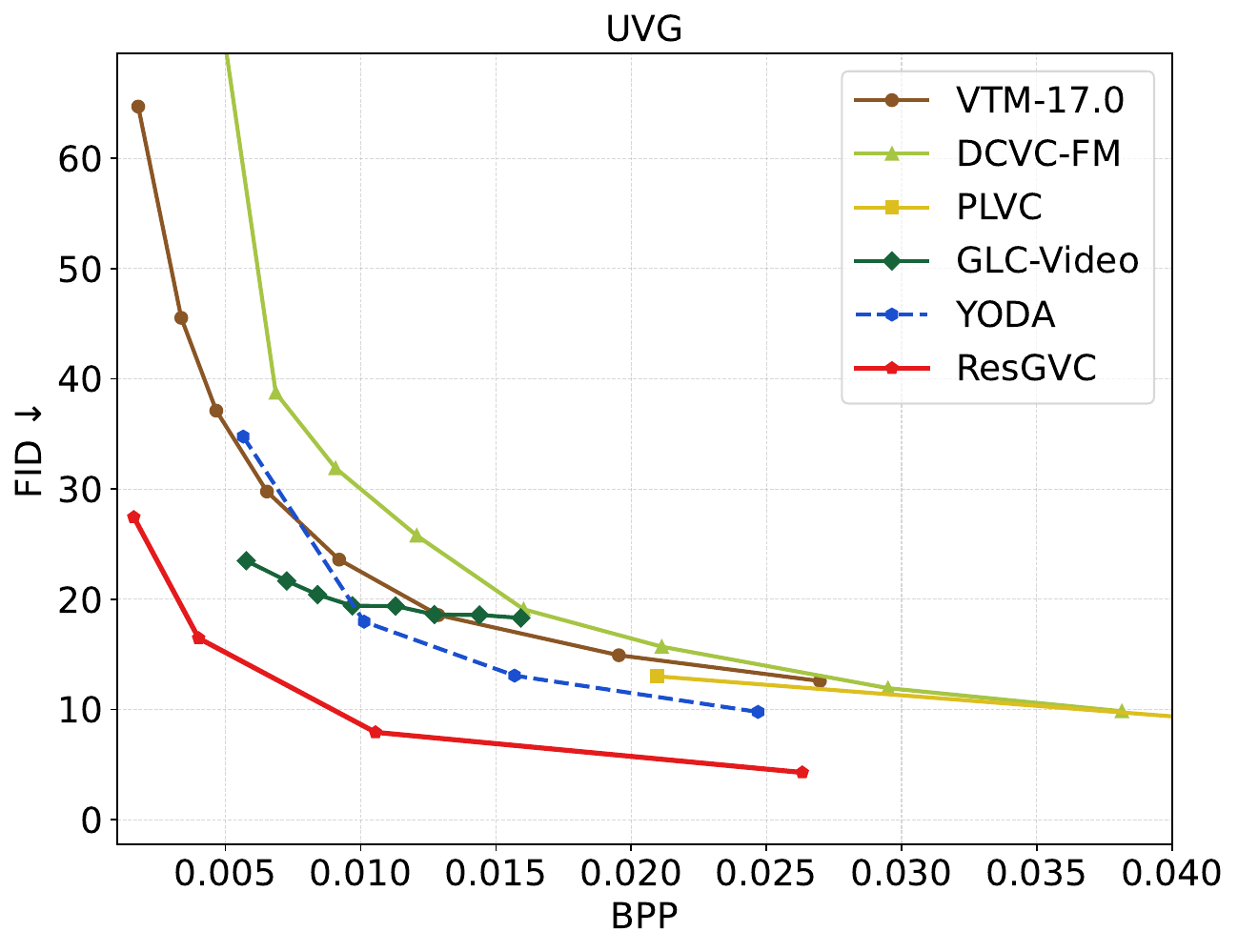}
    \label{fig:fid_UVG}
  \end{subfigure}
      \hfill
      \hspace{-10pt}
  \begin{subfigure}{0.25\linewidth}
    \includegraphics[width=\textwidth]{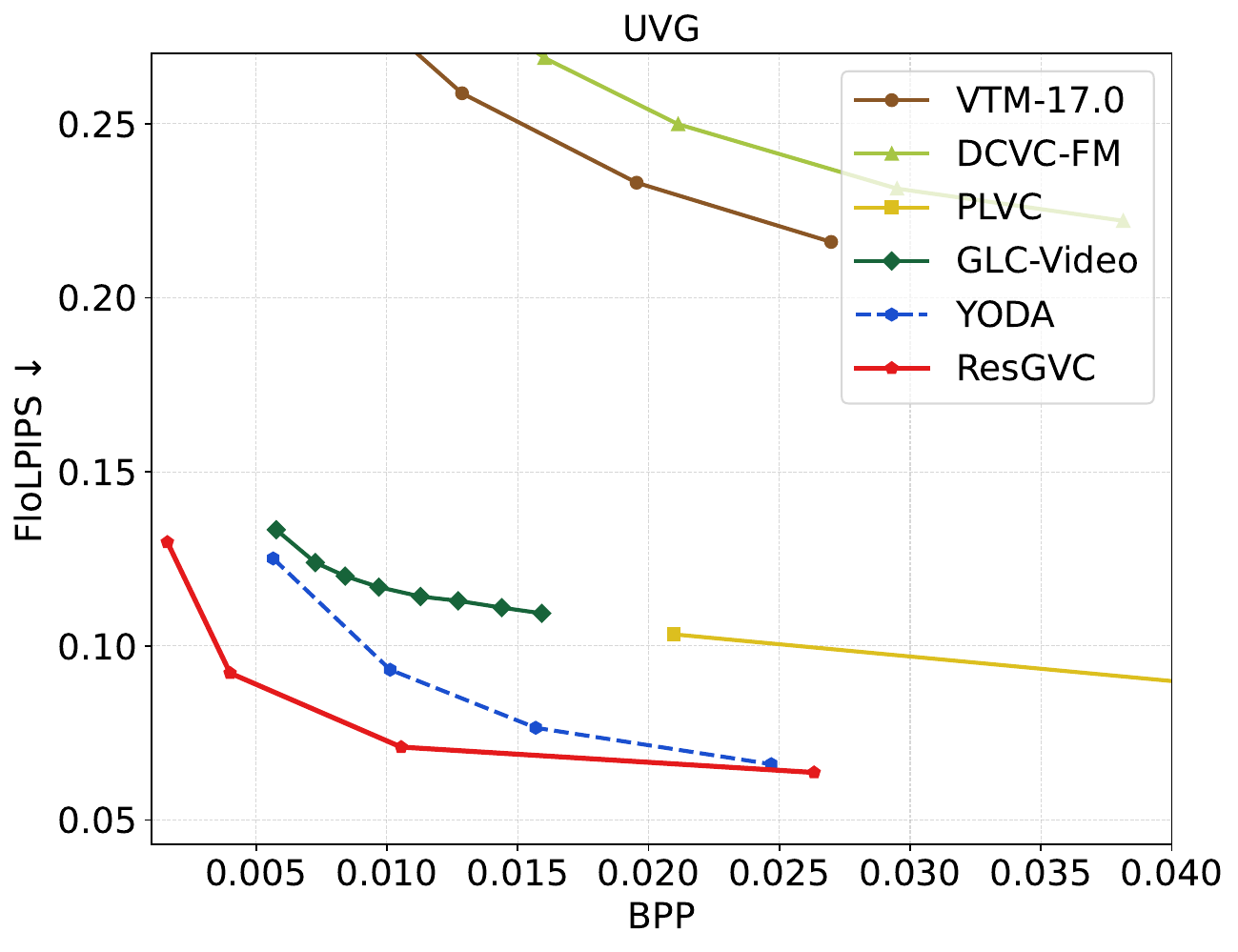}
    \label{fig:FloLPIPS_UVG}
  \end{subfigure}

  \vspace{-12pt}

  \begin{subfigure}{0.25\linewidth}
    \includegraphics[width=\textwidth]{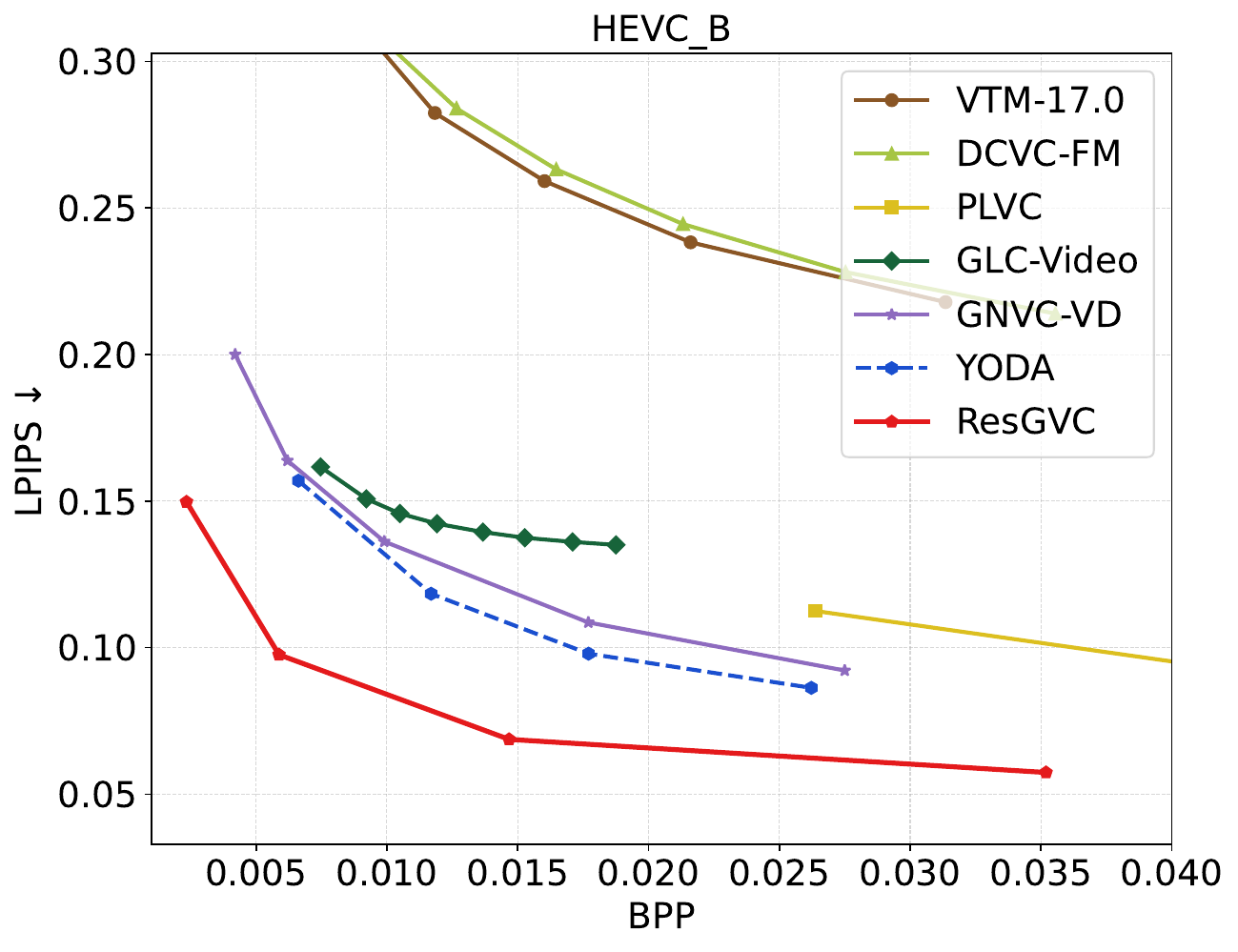}
    \label{fig:lpips_HEVC_B}
  \end{subfigure}
    \hfill
    \hspace{-10pt}
  \begin{subfigure}{0.25\linewidth}
    \includegraphics[width=\textwidth]{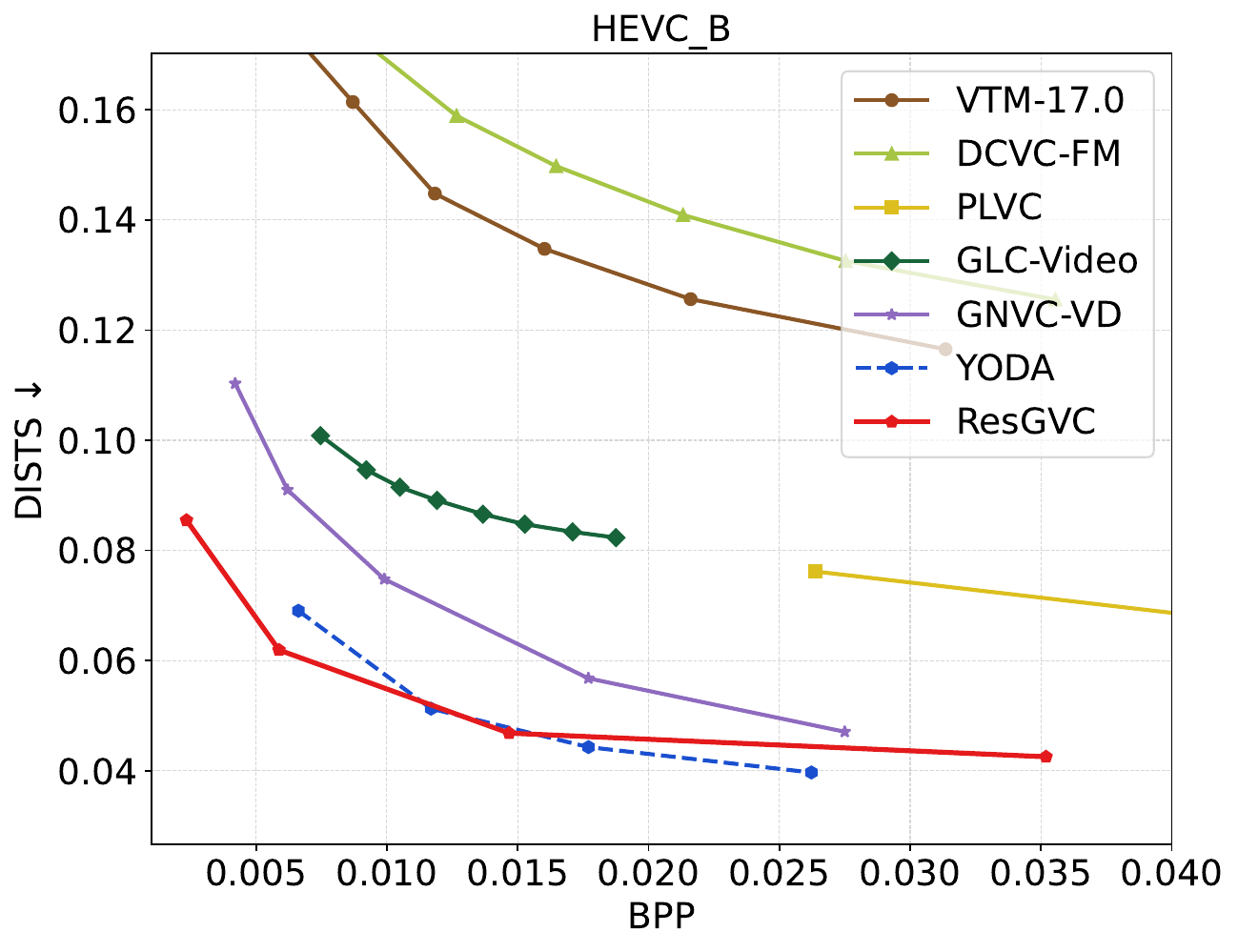}
    \label{fig:dist_HEVC_B}
  \end{subfigure}
      \hfill
      \hspace{-10pt}
  \begin{subfigure}{0.25\linewidth}
    \includegraphics[width=\textwidth]{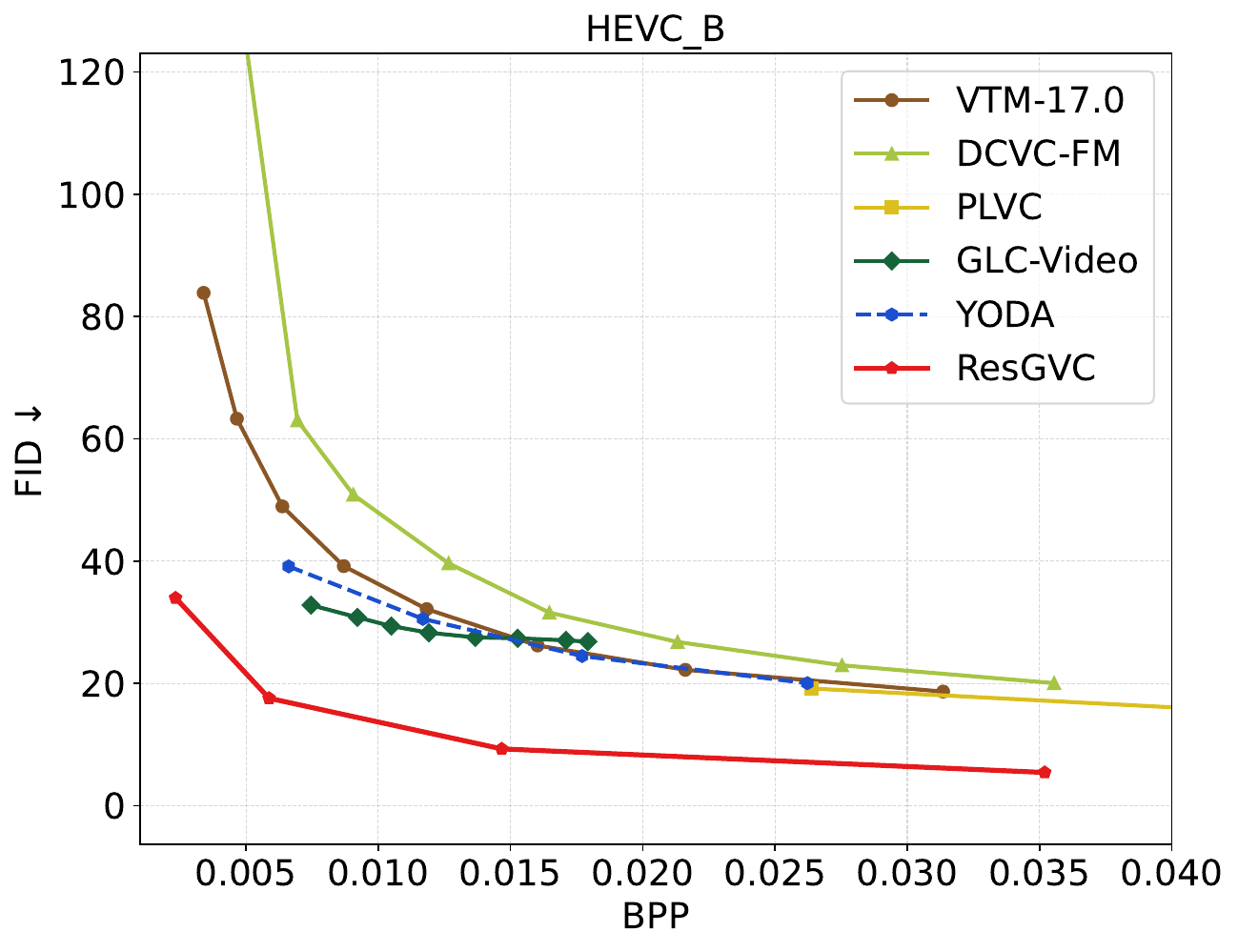}
    \label{fig:fid_HEVC_B}
  \end{subfigure}
      \hfill
      \hspace{-10pt}
  \begin{subfigure}{0.25\linewidth}
    \includegraphics[width=\textwidth]{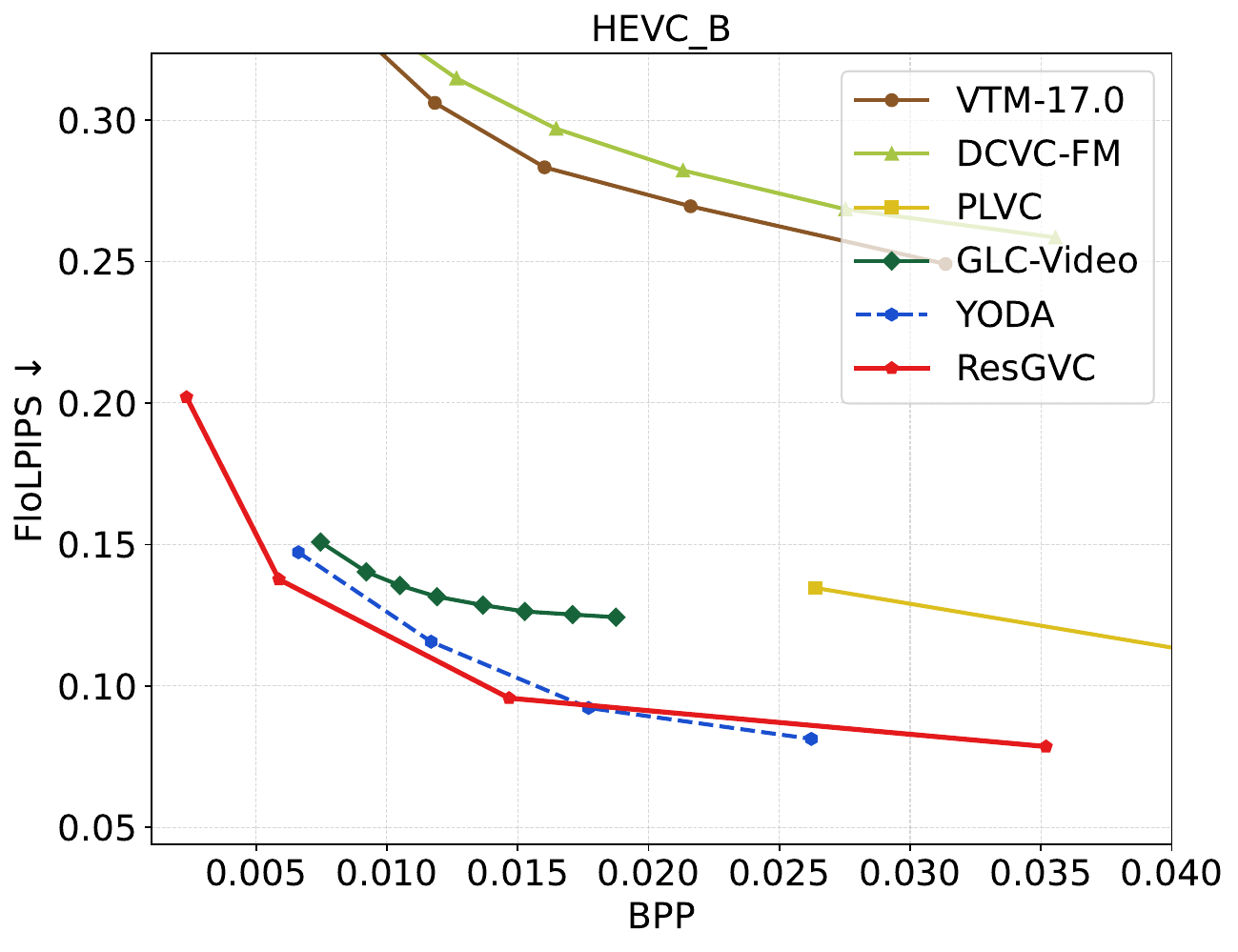}
    \label{fig:flolpips_HEVC_B}
  \end{subfigure}

  \vspace{-12pt}

  \begin{subfigure}{0.25\linewidth}
    \includegraphics[width=\textwidth]{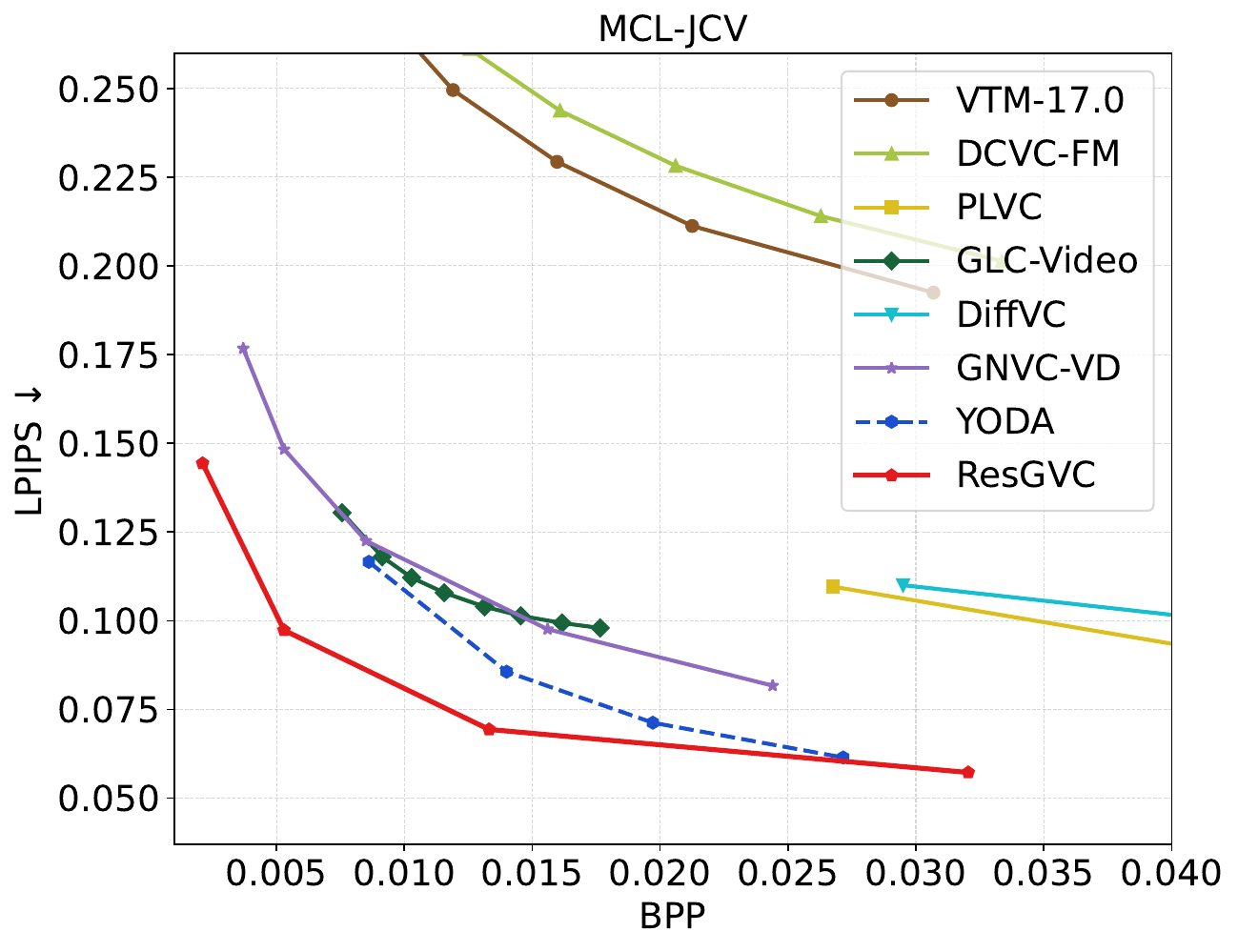}
    \label{fig:lpips_MCL-JCV}
  \end{subfigure}
    \hfill
    \hspace{-10pt}
  \begin{subfigure}{0.25\linewidth}
    \includegraphics[width=\textwidth]{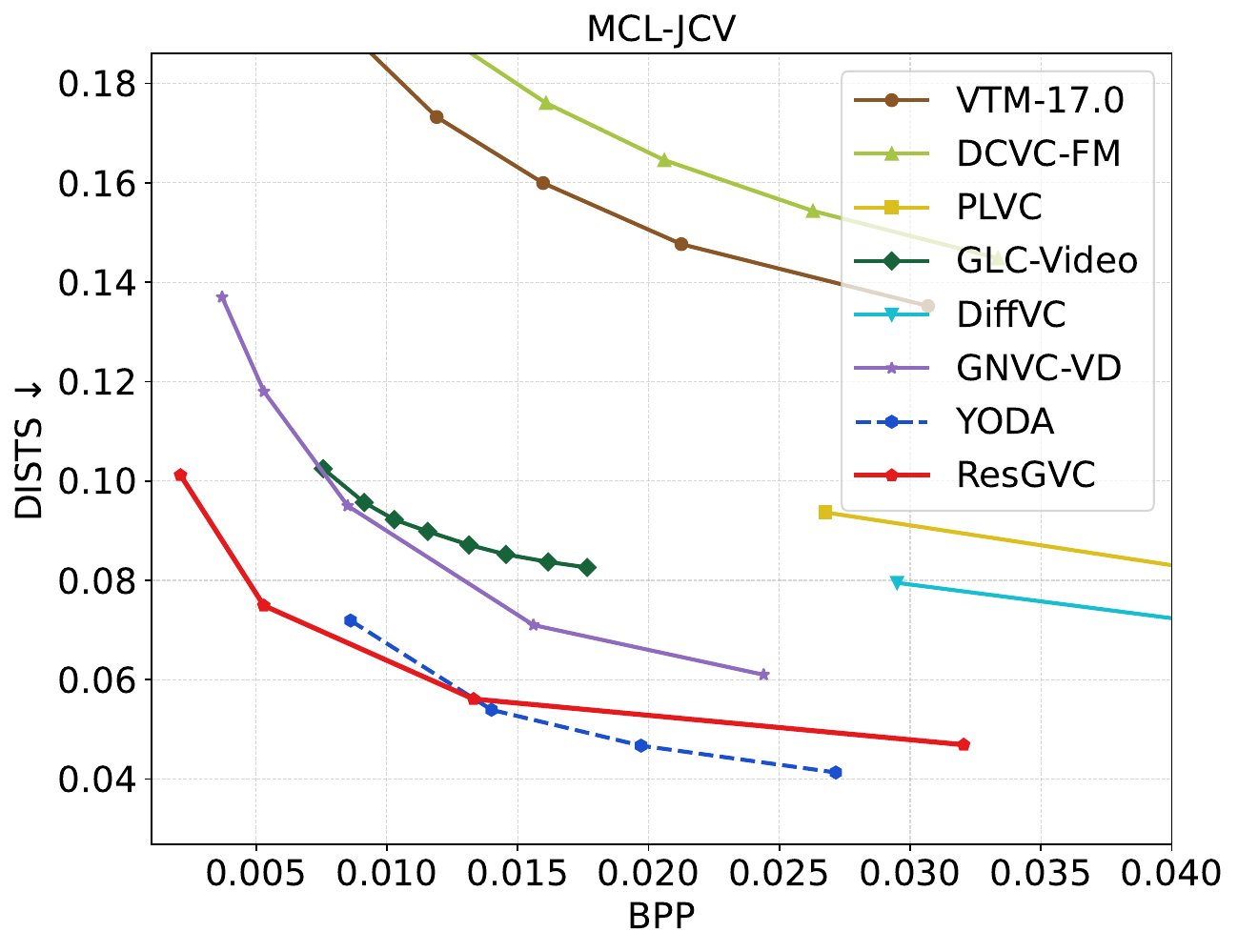}
    \label{fig:dist_MCL-JCV}
  \end{subfigure}
      \hfill
      \hspace{-10pt}
  \begin{subfigure}{0.25\linewidth}
    \includegraphics[width=\textwidth]{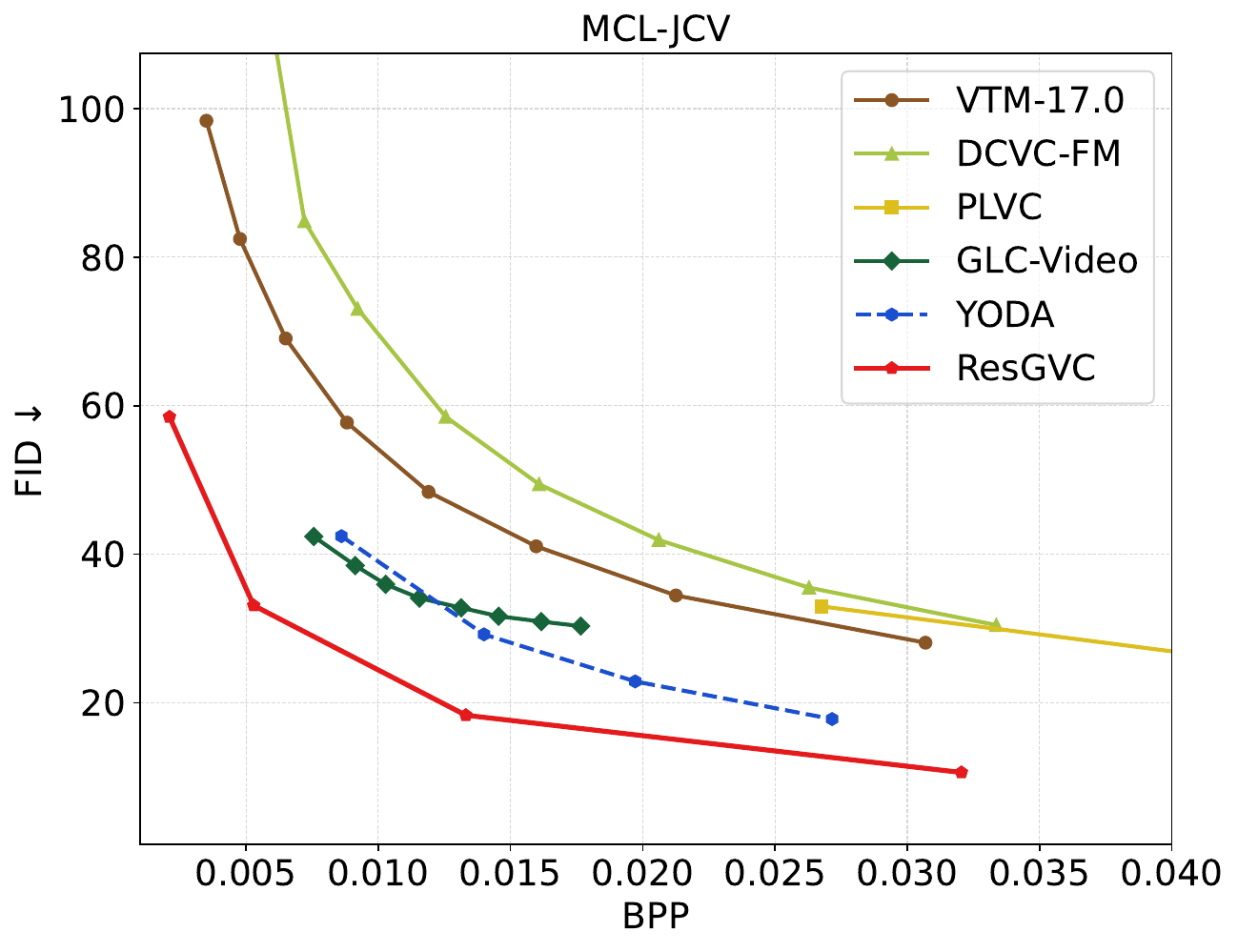}
    \label{fig:fid_MCL-JCV}
  \end{subfigure}
      \hfill
      \hspace{-10pt}
  \begin{subfigure}{0.25\linewidth}
    \includegraphics[width=\textwidth]{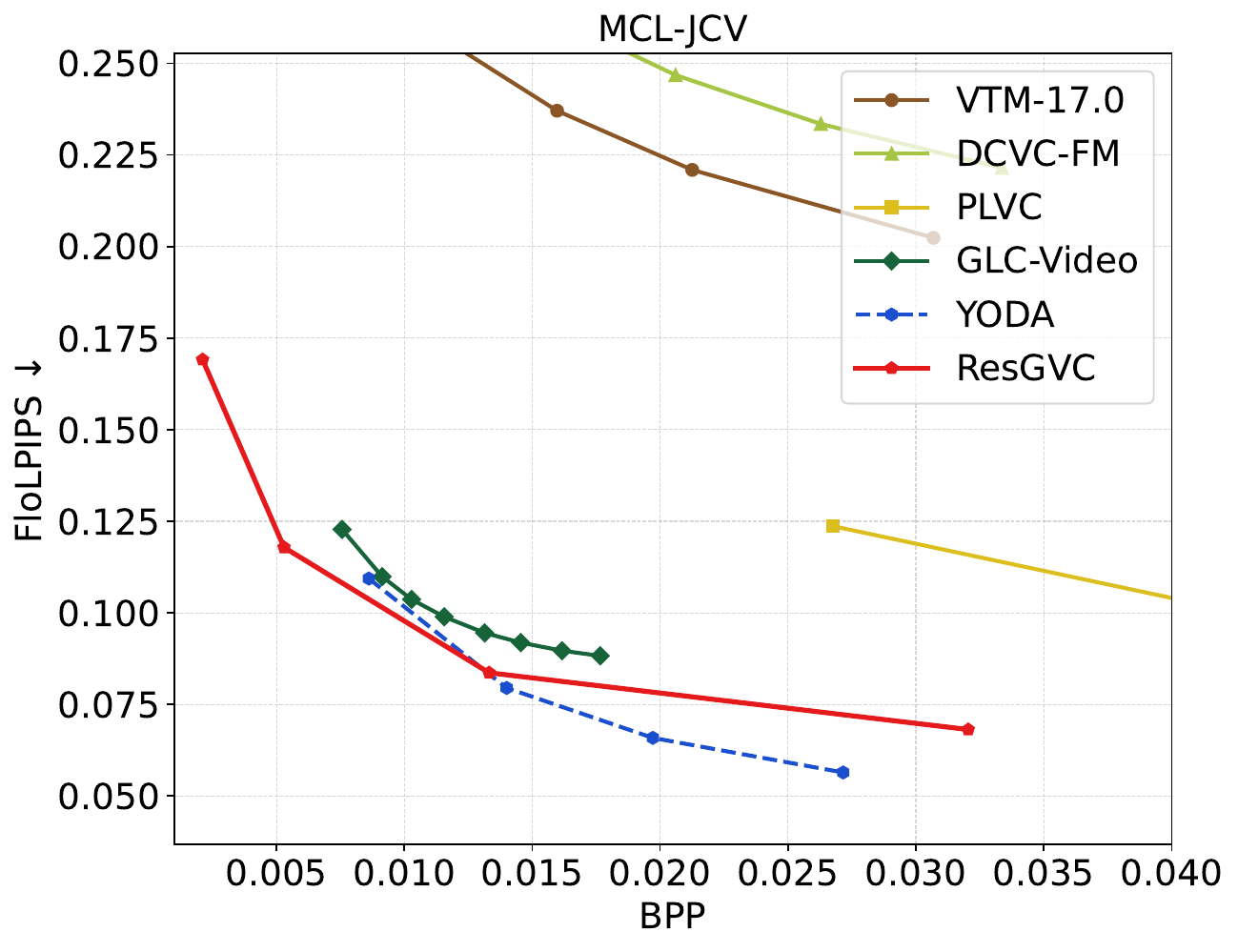}
    \label{fig:flolpips_MCL-JCV}
  \end{subfigure}
  \vspace{-20pt}
  \caption{Rate-perception curves on UVG, HEVC B, and MCL-JCV.}
  \label{fig:Rate-perception curve comparison}
  \vspace{-10pt}
\end{figure*}

\begin{table*}[t]
    \begin{center}
    \caption{\label{tab:comparison}
    BD-rate ($\%$) $\downarrow$ / BD-metric $\uparrow$ on UVG, HEVC B, and MCL-JCV. The best and second-best results are highlighted in \textcolor{red}{red} and \textcolor{blue}{blue}, respectively. ``N/A'' indicates that BD-rate cannot be computed due to insufficient overlap between the rate-metric curves. ``-'' denotes that the baseline does not report the result.
    }
    \vspace{-10pt}
    \renewcommand{\arraystretch}{1.25}
    \renewcommand{\tabcolsep}{0.5mm}
    \begin{adjustbox}{max width=\linewidth}
    \begin{tabular}{l|c|c|c|c||c|c|c|c||c|c|c|c}
    \hline
    \cline{1-13}
    & \multicolumn{4}{c||}{UVG} & \multicolumn{4}{c||}{HEVC B} & \multicolumn{4}{c}{MCL-JCV} \\
    \hline
    \cline{1-13}
    Method &
    LPIPS & DISTS & FID & FloLPIPS &
    LPIPS & DISTS & FID & FloLPIPS &
    LPIPS & DISTS & FID & FloLPIPS \\
    \cline{1-13}
    VTM-17.0
      & N/A / -0.136 & N/A / -0.092 & 21.5 / -3.0 & N/A / -0.177
      & N/A / -0.163 & N/A / -0.093 & 10.1 / -2.1 & N/A / -0.192
      & N/A / -0.150 & N/A / -0.109 & 89.6 / -14.3 & N/A / -0.163 \\
    \hline
    DCVC-FM
      & N/A / -0.145 & N/A / -0.108 & 66.9 / -8.8 & N/A / -0.203
      & N/A / -0.168 & N/A / -0.108 & 52.1 / -11.1 & N/A / -0.203
      & N/A / -0.165 & N/A / -0.126 & 143.5 / -24.3 & N/A / -0.188 \\
    \hline
    PLVC
      & 116.7 / -0.019 & 369.8 / -0.027 & 33.2 / -2.2 & 255.5 / -0.034
      & 91.3 / N/A & 594.2 / N/A & N/A / N/A & 200.5 / N/A
      & 208.1 / -0.048 & 658.7 / -0.052 & 129.7 / -14.9 & 319.4 / -0.067 \\
    \hline
    DiffVC
      & 425.5 / -0.070 & 395.8 / -0.036 & - / - & - / -
      & - / - & - / - & - / - & - / -
      & 275.1 / N/A & 382.3 / N/A & - / - & - / - \\
    \hline
    GLC-Video
      & 35.4 / -0.018 & N/A / -0.027 & \textcolor{blue}{-4.2} / \textcolor{blue}{0.4} & 56.2 / -0.022
      & 44.2 / -0.025 & N/A / -0.038 & \textcolor{blue}{-15.2} / \textcolor{blue}{1.6} & 33.1 / -0.019
      & 24.5 / -0.013 & N/A / -0.030 & 0.1 / -1.3 & 19.8 / -0.011 \\
    \hline
    GNVC-VD
      & 75.4 / -0.023 & 126.4 / -0.022 & - / - & - / -
      & 18.4 / -0.008 & 89.0 / -0.015 & - / - & - / -
      & 36.1 / -0.015 & 97.0 / -0.020 & - / - & - / - \\
    \hline
    YODA
      & \textcolor{blue}{0.0} / \textcolor{blue}{0.000} & \textcolor{blue}{0.0} / \textcolor{blue}{0.000} & 0.0 / 0.0 & \textcolor{blue}{0.0} / \textcolor{blue}{0.000}
      & \textcolor{blue}{0.0} / \textcolor{blue}{0.000} & \textcolor{blue}{0.0} / \textcolor{blue}{0.000} & 0.0 / 0.0 & \textcolor{blue}{0.0} / \textcolor{blue}{0.000}
      & \textcolor{blue}{0.0} / \textcolor{blue}{0.000} & \textcolor{blue}{0.0} / \textcolor{blue}{0.000} & \textcolor{blue}{0.0} / \textcolor{blue}{0.0} & \textcolor{red}{0.0} / \textcolor{red}{0.000} \\
    \hline
    \textbf{ResGVC}
      & \textcolor{red}{-74.9} / \textcolor{red}{0.031} & \textcolor{red}{-46.7} / \textcolor{red}{0.008} & \textcolor{red}{-68.4} / \textcolor{red}{10.2} & \textcolor{red}{-61.0} / \textcolor{red}{0.019}
      & \textcolor{red}{-68.1} / \textcolor{red}{0.042} & \textcolor{red}{-14.9} / \textcolor{red}{0.002} & \textcolor{red}{-78.9} / \textcolor{red}{18.6} & \textcolor{red}{-18.4} / \textcolor{red}{0.008}
      & \textcolor{red}{-48.3} / \textcolor{red}{0.016} & \textcolor{red}{-1.8} / \textcolor{red}{0.001} & \textcolor{red}{-54.4} / \textcolor{red}{11.2} & \textcolor{blue}{0.9} / \textcolor{blue}{-0.003} \\
    \hline
    \cline{1-13}
    \end{tabular}
    \end{adjustbox}
    \end{center}
    \vspace{-10pt}
\end{table*}

\subsection{Experimental Protocol}

\noindent{\textbf{Training details.}} \ours is trained on the OpenVid-1M dataset~\citep{openvid} with a rate-distortion loss function, \ie, $L = R + \lambda D$. $R$ denotes the total bitrate cost from the learned video codec and the proposed residual codec. $D$ denotes the reconstruction distortion, computed as the sum of L2 loss, LPIPS~\citep{lpips}, and DISTS~\citep{dists}. To evaluate compression performance at multiple bitrates, we train four independent models with trade-off parameter $\lambda$ sampled at four points by linear interpolation in logarithmic space over the range [0.025, 2.5]. Each model is trained with a GOP size of 8. $T$ is set to 5, indicating that five denoising steps are performed under the noise scheduler from~\citet{wan} to determine $\sigma_t$. The detailed training strategies are provided in the appendix.

\noindent{\textbf{Test details.}}
We evaluate \ours on the HEVC B~\citep{hevc}, UVG~\citep{uvg}, and MCL-JCV~\citep{mcl-jcv} datasets. Following~\citep{dcvc-dc, gnvc-vd},  raw YUV420 videos are converted to RGB using BT.709~\citep{dcvc-dc}, and the first 96 frames of each sequence are used for evaluation. Compression performance is measured by BD-rate and BD-metric~\citep{bd-rate}. Signal fidelity is measured by PSNR and MS-SSIM~\citep{ms-ssim}. Perceptual quality is assessed by LPIPS~\citep{lpips} and DISTS~\citep{dists}. To assess temporal perceptual consistency, we report FloLPIPS~\citep{flolpips}, a flow-guided LPIPS variant. Generative realism is measured by FID~\citep{fid}, following~\citep{hific}. Bitrate is reported in bits per pixel (BPP). \ours\ is evaluated with intra-period = -1 and GOP size = 8. Since \ours is trained with five denoising steps, we use two steps for residual coding and three steps for residual generation during testing, which gives the best overall compression performance. Analysis of this setting is provided in Section~\ref{subsec:ablation}.

\noindent{\textbf{Baselines.}} We compare \ours with three categories of baselines: 1)
the traditional video codec VTM-17.0~\citep{vvc}; 2) the fidelity-oriented NVC method DCVC-FM~\citep{dcvc-fm}; and 3) GVC methods, including PLVC~\citep{plvc}, GLC-Video~\citep{glc}, DiffVC~\citep{diffvc}, GNVC-VD~\citep{gnvc-vd}, and YODA~\citep{yoda}. For VTM-17.0, we follow the evaluation protocol from~\citet{dcvc-dc}. Since the code for DiffVC and GNVC-VD is publicly unavailable, we report the results from their original papers. Other baselines are tested with their default settings. We use the latest open-source method YODA as the anchor for BD-rate computation, rather than VTM-17.0 as usual, because the RD curves of VTM-17.0 are often far from those of GVC methods and overlap only marginally in quality range.

\subsection{Comparison Results}

\noindent{\textbf{Quantitative comparisons.}} Fig.~\ref{fig:Rate-perception curve comparison} shows the rate-perception curves on the three test sets. 
\begin{wrapfigure}{r}{0.5\linewidth}
  \vspace{-10pt}
  \centering
  \begin{subfigure}{0.5\linewidth}
    \includegraphics[width=\textwidth]{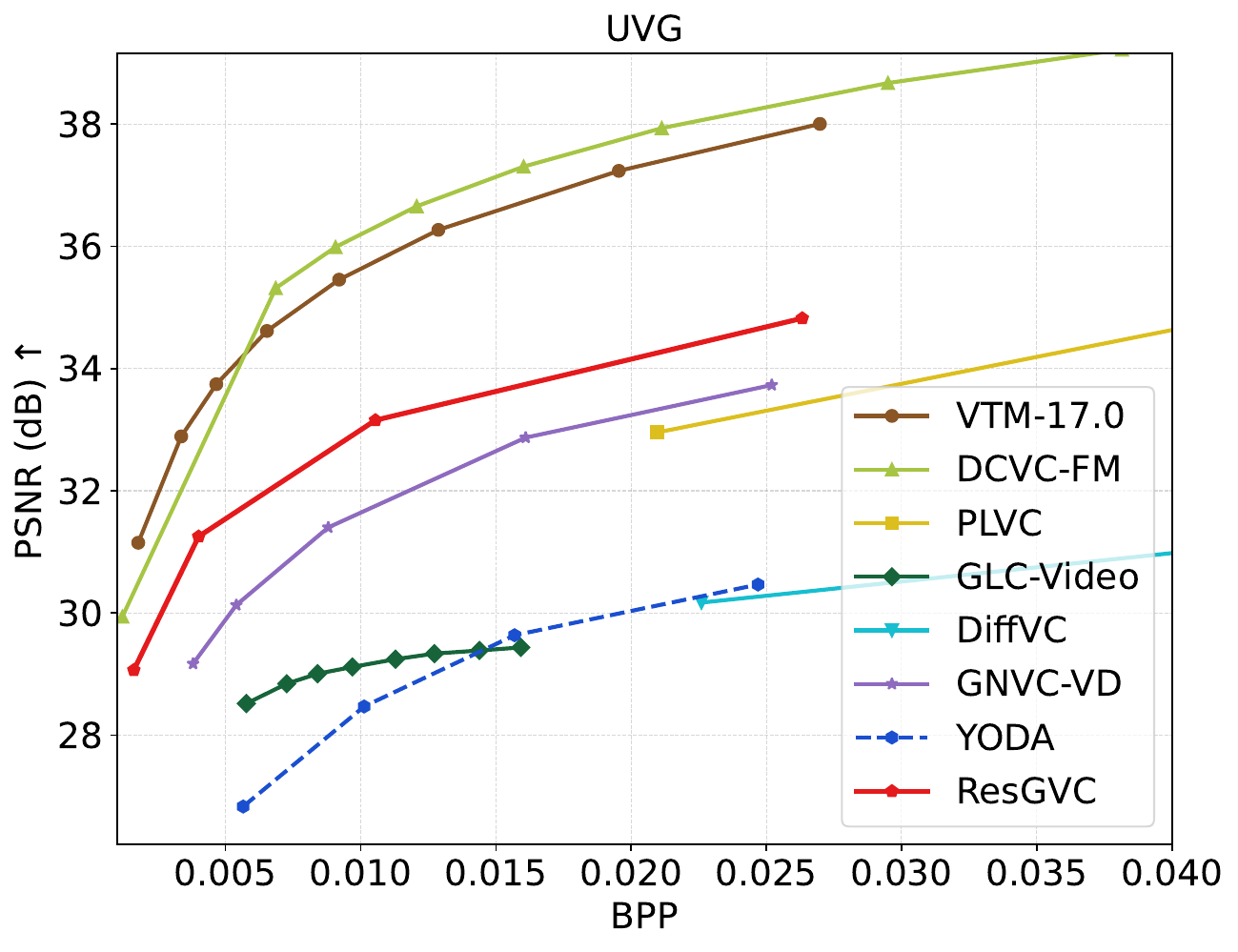}
    \label{fig:psnr_UVG}
  \end{subfigure}
      \hfill
  \hspace{-10pt}
  \begin{subfigure}{0.5\linewidth}
    \includegraphics[width=\textwidth]{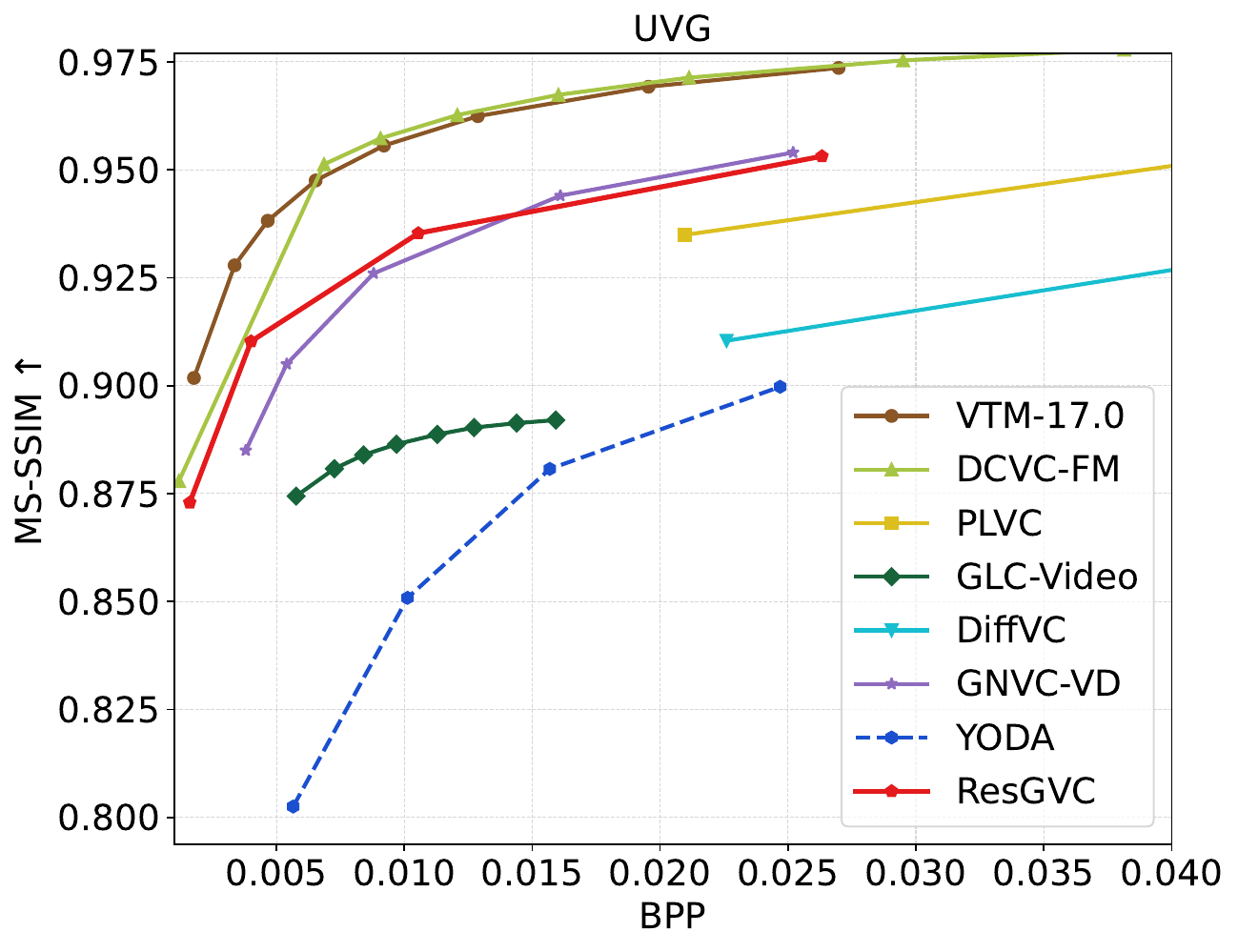}
    \label{fig:msssim_UVG}
  \end{subfigure}
  \\[-2pt]
  \vspace{-10pt}
  \begin{subfigure}{0.5\linewidth}
    \includegraphics[width=\textwidth]{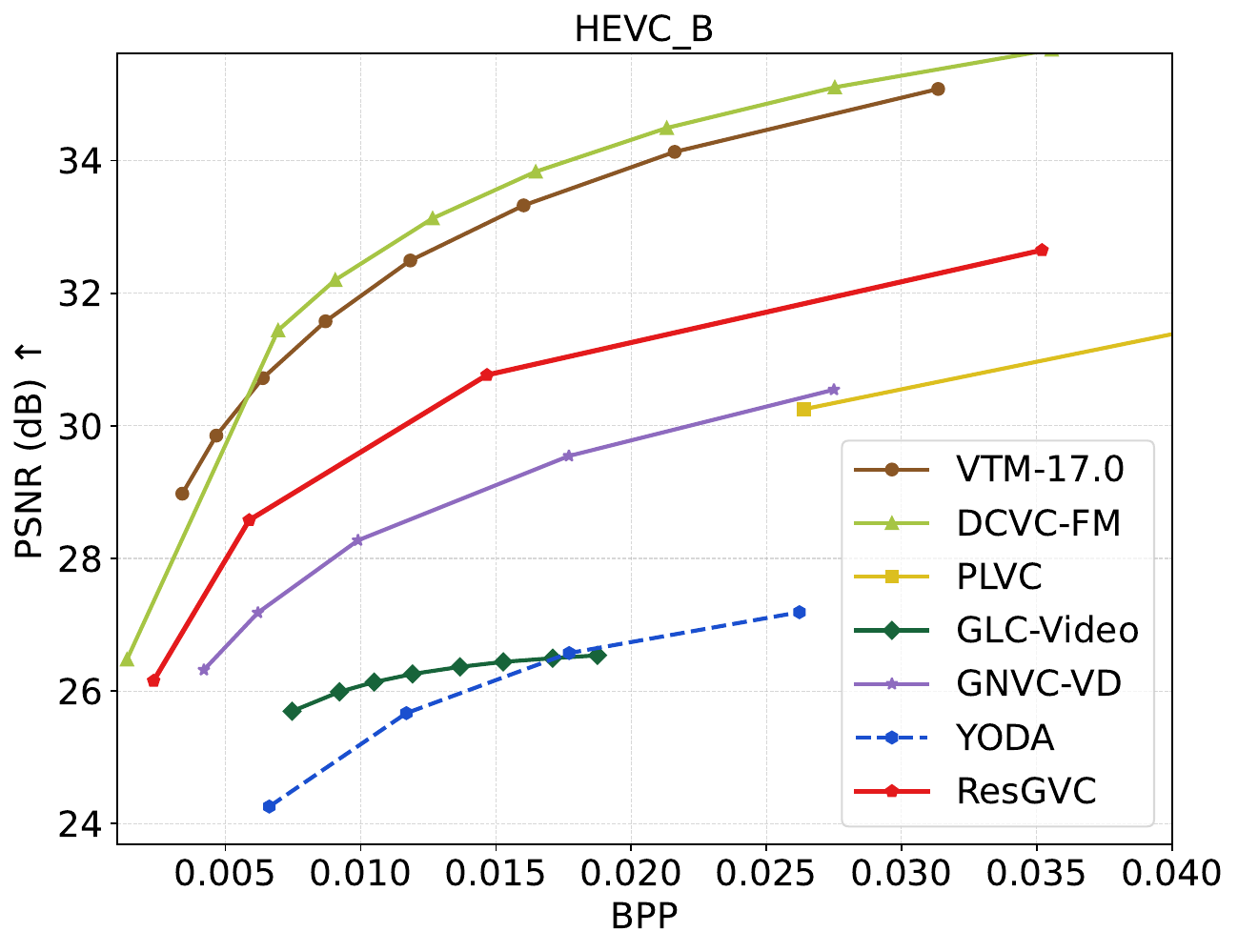}
    \label{fig:psnr_HEVC_B}
  \end{subfigure}
  \hfill
  \hspace{-10pt}
  \begin{subfigure}{0.5\linewidth}
    \includegraphics[width=\textwidth]{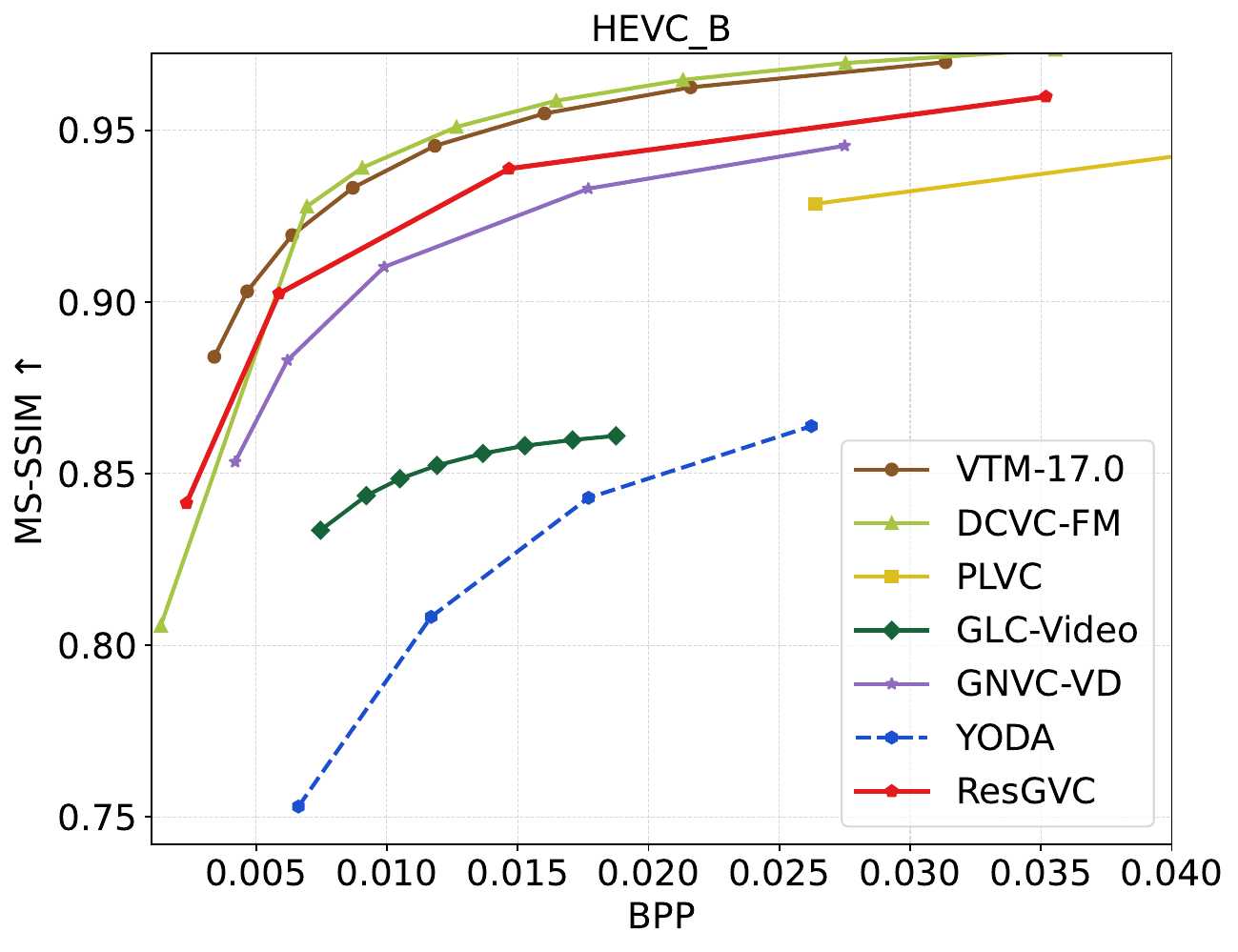}
    \label{fig:msssim_HEVC_B}
  \end{subfigure}
  \\[-2pt]
  \vspace{-10pt}
  \begin{subfigure}{0.5\linewidth}
    \includegraphics[width=\textwidth]{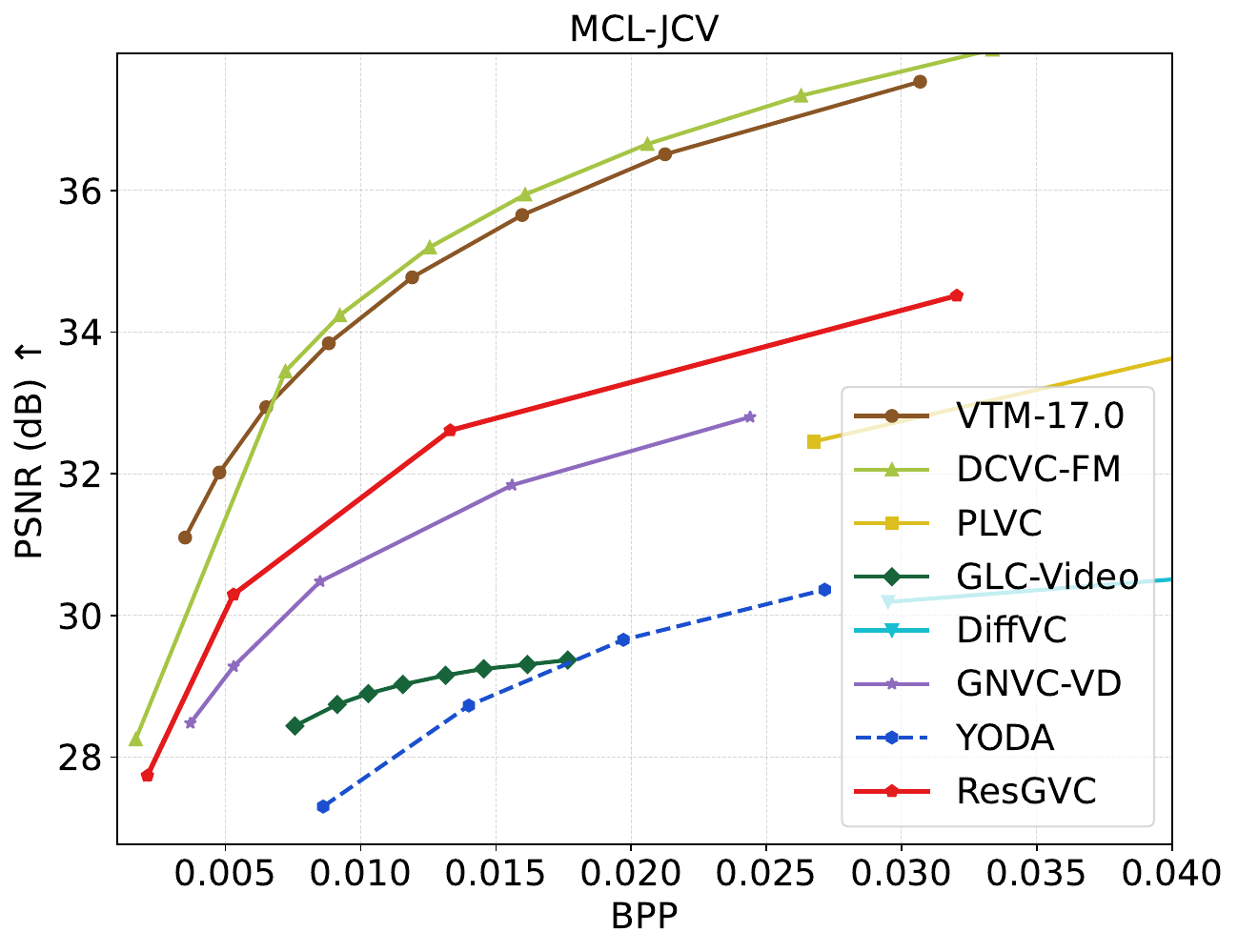}
    \label{fig:psnr_MCL-JCV}
  \end{subfigure}
  \hfill
  \hspace{-5pt}
  \begin{subfigure}{0.5\linewidth}
    \includegraphics[width=\textwidth]{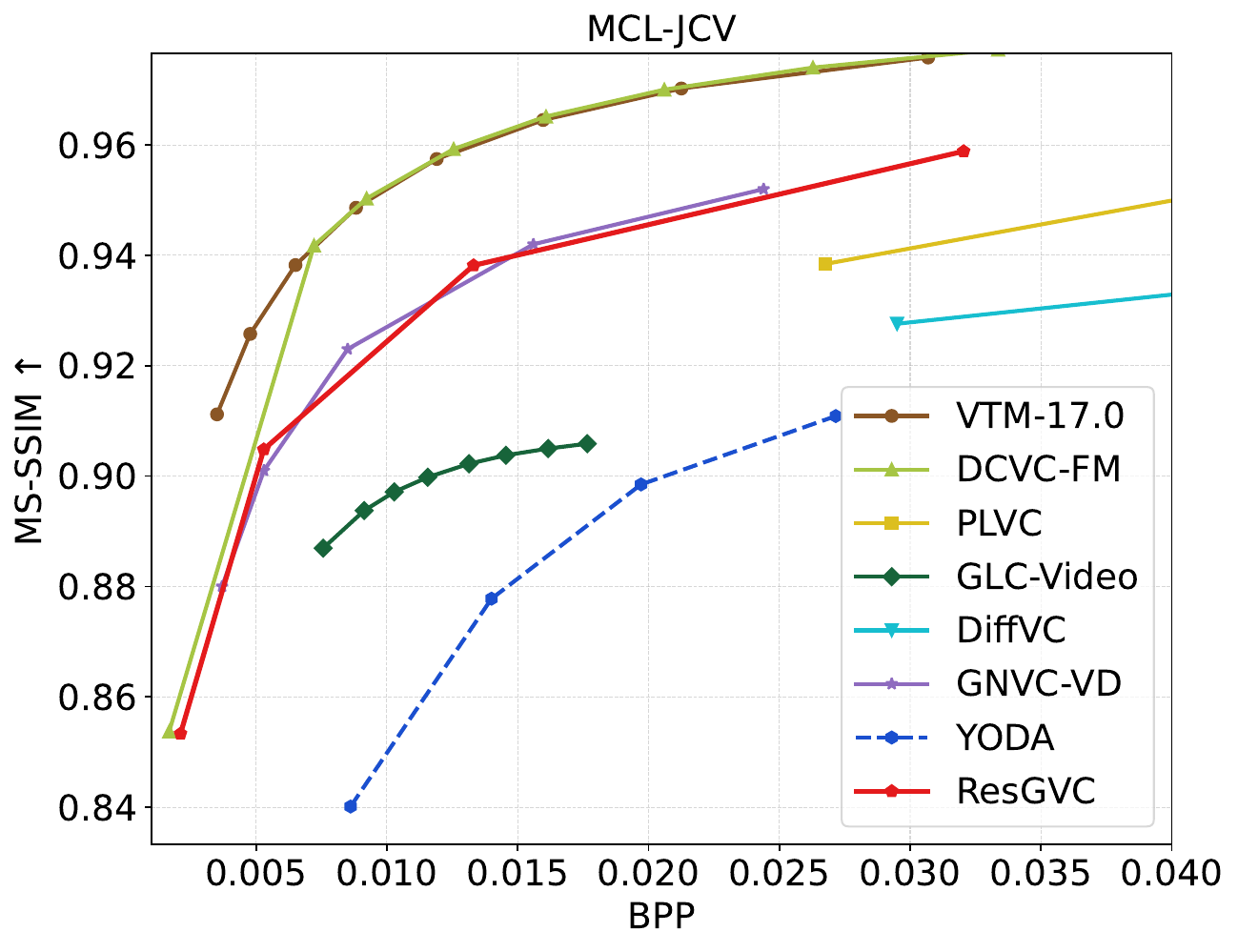}
    \label{fig:msssim_MCL-JCV}
  \end{subfigure}
  \vspace{-30pt}
  \caption{Rate-fidelity curves on UVG, HEVC B, and MCL-JCV.
  }
  \label{fig:Rate-fidelity curve comparison}
  \vspace{-10pt}
\end{wrapfigure}
The proposed \ours outperforms all baselines on most perceptual metrics, achieving notably lower LPIPS and FID at comparable bitrates. Table~\ref{tab:comparison} summarizes the BD-rate and BD-metric results. Averaged over the three test sets, \ours achieves bitrate savings of 63.7\% in LPIPS, 21.2\% in DISTS, 67.2\% in FID, and 26.1\% in FloLPIPS relative to the second-best method YODA. Additionally, as shown in Fig.~\ref{fig:Rate-fidelity curve comparison}, \ours delivers the best PSNR and MS-SSIM performance among all GVC methods, and is even comparable to the fidelity-oriented NVC method DCVC-FM at low bitrates. 
Notably, on the HEVC B test set, \ours surpasses DCVC-FM in MS-SSIM when the bitrate is below 0.005 BPP. Detailed BD-rate and BD-metric results for PSNR and MS-SSIM are provided in the appendix. Overall, these results show that \ours advances both perceptual quality and signal fidelity over previous GVC methods.

\begin{figure*}[t]
    \centering
    {\scriptsize
        \begin{tabularx}{\textwidth}{
          >{\centering\arraybackslash\hsize=2.15\hsize\linewidth=\hsize}X
          >{\centering\arraybackslash\hsize=0.77\hsize\linewidth=\hsize}X
          >{\centering\arraybackslash\hsize=0.77\hsize\linewidth=\hsize}X
          >{\centering\arraybackslash\hsize=0.77\hsize\linewidth=\hsize}X
          >{\centering\arraybackslash\hsize=0.77\hsize\linewidth=\hsize}X
          >{\centering\arraybackslash\hsize=0.77\hsize\linewidth=\hsize}X
        }
        & Original & DCVC-FM & GLC-Video & YODA & \textbf{\ours} \\
        \end{tabularx}
    }
    \vspace{-5pt}
    \includegraphics[width=\textwidth]{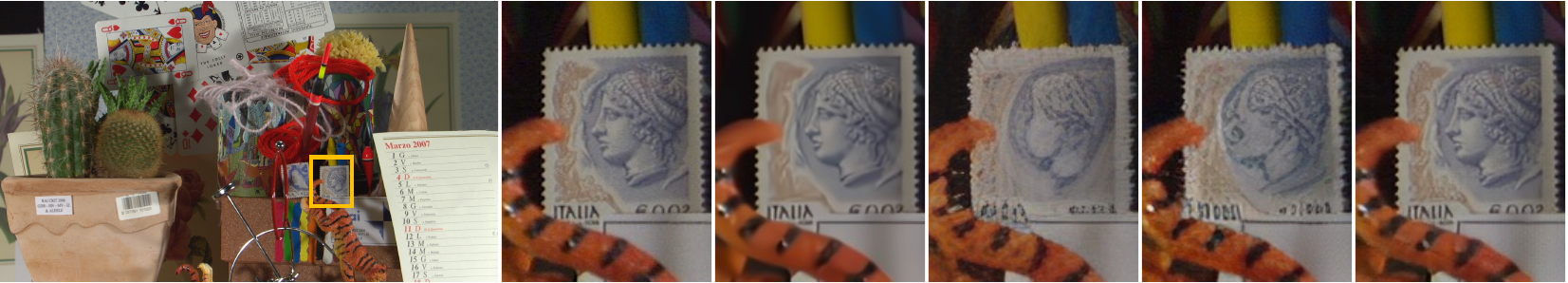}
    \vspace{5pt}
    {\scriptsize
        \begin{tabularx}{\textwidth}{
          >{\centering\arraybackslash\hsize=2.15\hsize\linewidth=\hsize}X
          >{\centering\arraybackslash\hsize=0.77\hsize\linewidth=\hsize}X
          >{\centering\arraybackslash\hsize=0.77\hsize\linewidth=\hsize}X
          >{\centering\arraybackslash\hsize=0.77\hsize\linewidth=\hsize}X
          >{\centering\arraybackslash\hsize=0.77\hsize\linewidth=\hsize}X
          >{\centering\arraybackslash\hsize=0.77\hsize\linewidth=\hsize}X
        }
        \textit{Cactus} & BPP$\downarrow$/LPIPS$\downarrow$ & 0.0158 / 0.262 & 0.0139 / 0.137 & 0.0154 / 0.097 & \textbf{0.0133} / \textbf{0.051} \\
        \end{tabularx}
    }
    \vspace{-20pt}
    \caption{Visual comparisons with open-source baselines on the \textit{Cactus} sequence from HEVC B.} 
    \label{fig:visualization on Cactus}
    \vspace{-15pt}
\end{figure*}

\noindent{\textbf{Qualitative comparisons.}} Fig.~\ref{fig:visualization on Cactus} presents a visual comparison of different methods. Owing to its optimization objective that prioritizes signal fidelity, DCVC-FM tends to produce blurred regions and fails to preserve fine-grained textures and structural details. Although GVC methods such as GLC-Video and YODA achieve higher perceptual quality than DCVC-FM, they often introduce spatial inconsistencies and hallucinated details, reflecting an overemphasis on perceptual quality at the expense of signal fidelity. By contrast, \ours more faithfully preserves object structures while maintaining rich textural details. These observations further suggest that, for GVC methods, signal fidelity should not be neglected when optimizing perceptual quality; otherwise, noticeable visual degradation and structural collapse may occur.

\noindent{\textbf{Complexity comparisons.}} We compare the per-frame encoding and decoding times of \ours with those of two recent methods, \ie, GNVC-VD and YODA. Among the GVC baselines, GNVC-VD delivers superior compression performance on fidelity-oriented metrics, whereas YODA excels on perceptual metrics. Since the official implementation of GNVC-VD is unavailable, we reproduce its model architecture solely for complexity evaluation, without training. As presented in Table~\ref{tab:complexity comparison}, \ours is faster than GNVC-VD in both encoding and decoding, mainly because GNVC-VD has a substantially larger model size. Compared with YODA, which uses single-step denoising, \ours has a slightly slower decoding speed due to its five-step denoising design. However, the compact model size of \ours helps offset the overhead introduced by the additional denoising steps, allowing it to maintain competitive decoding speed. Despite using fewer parameters, \ours still outperforms YODA in both perceptual quality and signal fidelity, further demonstrating the effectiveness of our pixel-space residual coding framework.

\begin{table*}[t]
\centering

\begin{minipage}[t]{0.475\textwidth}
\centering
\caption{Complexity comparison. ``Enc. T'' / ``Dec. T'' denotes the per-frame encoding / decoding time, tested on an NVIDIA H20 GPU.}
\vspace{-7pt}
\renewcommand{\arraystretch}{1.05}
\renewcommand{\tabcolsep}{2pt}
\begin{adjustbox}{max width=\linewidth}
\begin{tabular}{c|c|c|c}
\toprule
Method & Model Params (M) & Enc. T (ms) & Dec. T (ms) \\
\midrule
GNVC-VD & 2334.5 & 439  & 4225 \\
YODA & 1063.2 & 1139 & 729  \\
\textbf{Ours} & 593.9 & 346 & 772 \\
\bottomrule
\end{tabular}
\end{adjustbox}
\label{tab:complexity comparison}
\vspace{-8pt}
\end{minipage}
\hfill
\begin{minipage}[t]{0.5\textwidth}
\centering
\caption{BD-rate (\%)$\downarrow$ comparisons across various residual coding (RC) settings. The model with five-step RC in pixel space serves as the anchor.}
\vspace{-7pt}
\renewcommand{\arraystretch}{1.05}
\renewcommand{\tabcolsep}{3pt}
\begin{adjustbox}{max width=\linewidth}
\begin{tabular}{c|c|c|c|c}
\toprule
Method & LPIPS & DISTS & PSNR & FloLPIPS \\
\midrule
\textbf{w/ RC in pixel space} & 0.0 & 0.0 & 0.0 & 0.0 \\
w/o RC in pixel space & 44.6 & 59.1 & 12.1 & 34.2 \\
w/ RC in latent space & 17.4 & 21.3 & 28.6 & 15.1 \\
w/ CNN-based entropy model & 13.4 & 18.9 & 7.4 & 16.1 \\
\bottomrule
\end{tabular}
\end{adjustbox}
\label{tab:residual coding configuration}
\vspace{-8pt}
\end{minipage}

\vspace{-8pt}
\end{table*}

\subsection{Ablations}\label{subsec:ablation}

We conduct a series of ablation studies to verify the effect of key designs in the proposed \ours. All BD-rate results are averaged on three test sets.

\noindent{\textbf{Effect of pixel-space residual coding.}} \ours performs residual coding directly in pixel space. To validate this design, we conduct the ablations in Table~\ref{tab:residual coding configuration}, using the variant with five-step pixel-space residual coding as the anchor. Removing residual coding causes a substantial performance degradation, which confirms the necessity of residual coding. We then shift residual coding from pixel space to latent space by inserting \ours between the VAE encoder and decoder of Wan2.1~\citep{wan}, whose implementation details are provided in the appendix. This variant incurs 17.4\% and 28.6\% BD-rate losses in terms of LPIPS and PSNR, respectively, demonstrating that residual coding is more effective in pixel space than in latent space.

\noindent{\textbf{Effect of diffusion-based entropy model.}} Existing fidelity-oriented NVC and GVC methods typically rely on CNN-based entropy models to estimate the distributions used for entropy coding. In contrast, \ours adopts a diffusion-based entropy model, which leverages stronger generative priors for more accurate conditional probability estimation. As shown in Table~\ref{tab:residual coding configuration}, replacing our entropy model with the latest CNN-based design~\citep{dcvc-uf} degrades performance across all metrics, confirming the advantage of the diffusion-based entropy model for residual coding.

\begin{wraptable}{r}{0.5\linewidth}
\vspace{-10pt}
\caption{BD-rate (\%)$\downarrow$ comparison under different skip settings for residual coding. ``RC'' and ``RG'' denote residual coding and residual generation, respectively.}
\centering
\vspace{-7.5pt}
\renewcommand{\arraystretch}{1.05}
\renewcommand{\tabcolsep}{3pt}
\begin{adjustbox}{max width=\linewidth}
    \begin{tabular}{c|c|c|c|c}
    \toprule
    Method & LPIPS & DISTS & PSNR & FloLPIPS \\
    \midrule
    0-step RC + 5-step RG & 11.4 & 11.6 & 9.2 & 8.8 \\
    1-step RC + 4-step RG & 7.7 & 8.2 & 7.1 & 6.5 \\
    \textbf{2-step RC + 3-step RG (Ours)} & 0.0 & 0.0 & 0.0 & 0.0 \\
    3-step RC + 2-step RG & 2.4 & 4.5 & 1.8 & 2.7 \\
    4-step RC + 1-step RG & 5.4 & 5.8 & 3.5 & 3.6 \\
    5-step RC + 0-step RG & 6.7 & 7.3 & 3.4 & 4.2 \\
    \bottomrule
    \end{tabular}
\end{adjustbox}
\label{tab:different steps for residual coding and residual generation}
\vspace{-10pt}
\end{wraptable}

\noindent{\textbf{Effect of the skip mode for residual coding.}} In \ours, residual coding can be skipped at selected denoising steps, where residual segments are instead generated at the decoder by the diffusion model. This skip mode enables flexible switching between residual coding and residual generation at different denoising steps to optimize rate-distortion performance. As presented in Table~\ref{tab:different steps for residual coding and residual generation}, we fix the total number of denoising steps to five and vary the number of residual-coding steps from zero to five, with the remaining steps handled by residual generation. The best performance is achieved with two coding steps and three generation steps, indicating that the most fidelity-critical residual segments are transmitted in the first two steps. Using more coding steps degrades performance by spending additional bitrate on residuals that can be reliably predicted from generative priors at the decoder, while using fewer coding steps weakens source constraints and thus reduces reconstruction quality.

\section{Conclusion}
In this paper, we present \ours, a residual coding framework for generative video compression that guides diffusion denoising toward the original pictures by transmitting residual signals. \ours first encodes the original pictures with a learned video codec, and then decomposes the distortion between the codec reconstruction and the original pictures into a sequence of residual segments over denoising steps. These residual segments can further by compressed by a novel residual codec with a diffusion transformer as an entropy model, enabling explicit source constraints to be introduced throughout denoising rather than relying on decoder-side generation alone. By denoising directly in pixel space, \ours also avoids the irreversible information loss caused by latent representation bottlenecks. Experimental results demonstrate that \ours achieves state-of-the-art compression performance, reducing BD-rate by 63.7\% in LPIPS and 83.8\% in PSNR compared with the previous state-of-the-art method, while also delivering superior visual quality.

\subsection*{AI use statement}

In this work, we used GPT-5.4 to improve the clarity, grammar, and formal tone of the writing in all sections, and to retrieve and translate related work in support of a more comprehensive understanding of this research field. We did not use generative AI tools to help develop theoretical models or conceptual frameworks, formulate mathematical claims, provide critical ingredients for proving mathematical claims, design or provide feedback on research methodology or experiments, implement methods, clean or reformat datasets, support qualitative and thematic data analysis, or interpret results. The generation of synthetic datasets is not applicable to this work. All AI-generated suggestions were reviewed and edited by the authors. We take full responsibility for the final content of this work, including any text, claims, or artifacts produced with the assistance of generative AI.

\bibliography{main}
\bibliographystyle{main}

\newpage

\appendix
\section{Appendix}
\subsection{Model Implementations}
\noindent{\textbf{Analysis Transform and Synthesis Transform.}} The proposed residual codec employs analysis transform $g_\mathrm{a}$ and synthesis transform $g_\mathrm{s}$ to reduce the redundancy within each residual segment. $g_\mathrm{a}$ and $g_\mathrm{s}$ are built upon the encoder and decoder architecture of the inter model in DCVC-UF HT-L~\citep{dcvc-uf}, with two principal distinctions. First, our $g_\mathrm{a}$ and $g_\mathrm{s}$ are shared by both intra-coding and inter-coding, with the adapter introduced for the intra-frame to align the input channels of $g_\mathrm{a}$. By contrast, DCVC-UF employs two separate models for intra- and inter-coding. Second, we employ the previously reconstructed GOP $\hat{\mathbf{x}}_{0}^{\mathrm{p}}$ as condition information, whereas DCVC-UF relies on memory updated using features from decoded frames.

\noindent{\textbf{Diffusion-based Entropy Model.}} In our framework, the video pixel diffusion transformer (VideoPiT) serves as an entropy model, in which each PiT block is built upon the T2I variant of PixelDiT~\citep{pixeldit}. We employ the previously reconstructed GOP $\hat{\mathbf{x}}_{0}^{\mathrm{p}}$ as the conditioning input, whereas PixelDiT conditions on features derived from the text embedding. In VideoPiT, the input $\hat{\mathbf{x}}_{t}$ first undergoes $16 \times 16$ patchification, followed by six PiT blocks. The output feature of the final PiT block is linearly projected to produce the probability parameters $\boldsymbol{\mu}_t$ and $\boldsymbol{\sigma}_t$ for entropy coding. Notably, both $\boldsymbol{\mu}_t$ and $\boldsymbol{\sigma}_t$ have spatial dimensions downsampled by a factor of 16 relative to $\hat{\mathbf{x}}_{t}$.

\subsection{More Training Details}

We initialize the learned video codec~\citep{dcvc-uf} with its pretrained weights, and the proposed residual codec is trained from scratch. AdamW~\citep{adamw} is used as the optimizer. For the training set OpenVid-1M~\citep{openvid}, we resize and crop each training sample to $448\times 448$. Adopting the noise schedule from Wan2.1~\citep{wan}, the noise levels $\sigma_t$ from $t=5$ down to $t=0$ are configured as [0.83, 0.77, 0.68, 0.56, 0.36, 0.0], where the initial magnitude of 0.83 reflects our assumption that the decoded video from the learned video codec exhibits a noise (distortion) ratio of 0.83. The loss function is formulated as:
\begin{equation}
    L = R_\mathrm{LVC} + R_\mathrm{ResC} + \lambda (\| \mathbf{x}_0 - \hat{\mathbf{x}}_0 \|_2^2 + L_\mathrm{LPIPS}(\mathbf{x}_0, \hat{\mathbf{x}}_0) + L_\mathrm{DISTS}(\mathbf{x}_0, \hat{\mathbf{x}}_0)),
\end{equation}
where $R_\mathrm{LVC}$ denotes the bitrate cost from the learned video codec and $R_\mathrm{ResC}$ denotes the bitrate cost from the proposed residual codec. $L_\mathrm{LPIPS}$ and $L_\mathrm{DISTS}$ represent the perceptual losses measured by LPIPS~\citep{lpips} and DISTS~\citep{dists}. $\mathbf{x}_0$ and $\hat{\mathbf{x}}_0$ denote the original pictures and the reconstruction.

To stabilize model optimization, we adopt a multi-stage training strategy as follows. All training is conducted on eight NVIDIA H20 GPUs.

\noindent{\textbf{Stage 1.}} We train only the residual codec with single-step denoising, where $\sigma_1=0.83$  and $\sigma_0=0$. The trade-off parameter $\lambda$ is set to 8. The number of training frames is progressively increased from 2 to 33, while the batch size is reduced from 256 to 16 accordingly. The learning rate is set to $5\times10^{-5}$.

\noindent{\textbf{Stage 2.}} We jointly train the learned video codec and the proposed residual codec. $\lambda$ is sampled at four points using linear interpolation in logarithmic space over the range [0.025, 2.5], resulting in four independent models for evaluating compression performance at different bitrates. The number of training frames is set to 33, the batch size to 8, and the learning rate to $5\times10^{-5}$.

\noindent{\textbf{Stage 3.}} We progressively scale the number of denoising steps, \ie, $T$, from 1 to 5. Standard flow matching typically samples a random step $t$ during each training iteration and optimizes the objective in Eq.~\ref{eq:flow matching loss} under a single-step denoising paradigm, akin to the setting adopted in Stage 1. However, given our multi-step residual coding design, compression distortions from the decoded residuals propagate across successive steps. Consequently, we employ multi-step denoising during training to adapt the model to such error propagation. To ensure stable optimization, we only backpropagate the gradient at step 0. The number of training frames is set to 33, the batch size to 8, and the learning rate to $2\times10^{-5}$.

\noindent{\textbf{Stage 4.}} We increase the number of training frames to 97. All other training configurations remain the same as in Stage 3.

\noindent{\textbf{Stage 5.}} During multi-step residual coding, we allow gradients to backpropagate across steps. The number of training frames is set to 97, the batch size to 4, and the learning rate to $1\times10^{-5}$.

\subsection{Further Comparison Results}
\begin{table*}[t]
    \begin{center}
    \caption{\label{tab:comparison_fidelity_oriented_metrics}
    BD-rate ($\%$) $\downarrow$ / BD-metric $\uparrow$ for PSNR and MS-SSIM on UVG, HEVC B, and MCL-JCV, using YODA as the anchor. The best and second-best results are highlighted in \textcolor{red}{red} and \textcolor{blue}{blue}, respectively. ``N/A'' indicates that BD-rate cannot be computed due to insufficient overlap between the rate--metric curves. ``-'' denotes that the baseline does not report the result.
    }
    \vspace{-7.5pt}
    \renewcommand{\arraystretch}{1.25}
    \renewcommand{\tabcolsep}{1.5mm}
    \begin{adjustbox}{max width=0.75\linewidth}
    \begin{tabular}{l|c|c||c|c||c|c}
    \hline
    \cline{1-7}
    & \multicolumn{2}{c||}{UVG} & \multicolumn{2}{c||}{HEVC B} & \multicolumn{2}{c}{MCL-JCV} \\
    \hline
    \cline{1-7}
    Method & PSNR & MS-SSIM & PSNR & MS-SSIM & PSNR & MS-SSIM \\
    \cline{1-7}
    VTM-17.0
      & \textcolor{blue}{N/A} / \textcolor{blue}{7.221} & \textcolor{blue}{N/A} / \textcolor{blue}{0.101}
      & \textcolor{blue}{N/A} / \textcolor{blue}{6.886} & \textcolor{blue}{N/A} / \textcolor{blue}{0.132}
      & \textcolor{blue}{N/A} / \textcolor{blue}{6.582} & \textcolor{blue}{N/A} / \textcolor{blue}{0.082} \\
    \hline
    DCVC-FM
      & \textcolor{red}{-94.0} / \textcolor{red}{7.766} & \textcolor{red}{-93.1} / \textcolor{red}{0.103}
      & \textcolor{red}{-92.7} / \textcolor{red}{7.334} & \textcolor{red}{-89.5} / \textcolor{red}{0.135}
      & \textcolor{red}{-87.3} / \textcolor{red}{6.834} & \textcolor{red}{-84.9} / \textcolor{red}{0.083} \\
    \hline
    PLVC
      & N/A / 2.848 & N/A / 0.040
      & N/A / N/A & N/A / N/A
      & N/A / 2.134 & N/A / 0.028 \\
    \hline
    DiffVC
      & 9.5 / -0.158 & N/A / 0.014
      & - / - & - / -
      & 20.9 / N/A & N/A / N/A \\
    \hline
    GLC-Video
      & -29.1 / 0.760 & -48.8 / 0.040
      & -24.7 / 0.530 & -44.9 / 0.042
      & -28.2 / 0.710 & -45.4 / 0.032 \\
    \hline
    GNVC-VD
      & -72.3 / 3.265 & -79.0 / 0.075
      & -74.2 / 3.020 & -81.0 / 0.105
      & -69.5 / 2.876 & -74.7 / 0.061 \\
    \hline
    YODA
      & 0.0 / 0.000 & 0.0 / 0.000
      & 0.0 / 0.000 & 0.0 / 0.000
      & 0.0 / 0.000 & 0.0 / 0.000 \\
    \hline
    \textbf{ResGVC}
      & -87.6 / 4.535 & -88.1 / 0.079
      & -85.1 / 4.625 & -87.4 / 0.118
      & -78.9 / 3.977 & -77.8 / 0.060 \\
    \hline
    \cline{1-7}
    \end{tabular}
    \end{adjustbox}
    \end{center}
    \vspace{-5pt}
\end{table*}

We report the BD-rate and BD-metric results for fidelity-oriented measures, \ie, PSNR and MS-SSIM, in Table~\ref{tab:comparison_fidelity_oriented_metrics}. Among all methods, the fidelity-oriented NVC approach DCVC-FM achieves the best performance, while the traditional video codec VTM-17.0 delivers the second-best results. This is expected, as both are optimized to prioritize signal fidelity rather than perceptual quality. Among the GVC methods, the proposed \ours demonstrates clear superiority over the other baselines, achieving approximately 10\% bitrate savings on each test set compared with the second-best method GNVC-VD. Notably, the performance gap between DCVC-FM and \ours is relatively small, with the BD-rate difference remaining below 10\% across all three test sets.

\begin{table}[t]
\caption{BD-rate (\%)$\downarrow$ comparison of performing residual coding (RC) and residual generation (RG) in latent space \textit{vs.} pixel space.}
\centering
\vspace{-7.5pt}
\renewcommand{\arraystretch}{1.05}
\renewcommand{\tabcolsep}{5pt}
\begin{adjustbox}{max width=0.75\linewidth}
    \begin{tabular}{c|c|c|c|c|c}
    \toprule
    & Method & LPIPS & DISTS & PSNR & FloLPIPS \\
    \midrule
    (a) & 5-step RC in latent space & 15.4 & 17.5 & 30.8 & 10.3 \\
    (b) & 5-step RC in pixel space & 6.7 & 7.3 & 3.4 & 4.2 \\
    \midrule
    (c) & 5-step RG in latent space & 7.4 & 8.1 & 22.6 & 4.6 \\
    (d) & 5-step RG in pixel space & 11.4 & 11.6 & 9.2 & 8.8 \\
    \midrule
    (e) & 2-step RC and 3-step RG in latent space & 13.2 & 13.1 & 24.7 & 5.6 \\
    (f) &\textbf{2-step RC and 3-step RG in pixel space (Ours)} & \textbf{0.0} & \textbf{0.0} & \textbf{0.0} & \textbf{0.0} \\   
    \bottomrule
    \end{tabular}
\end{adjustbox}
\label{tab:residual coding and residual generation in latent space or pixel space}
\vspace{-5pt}
\end{table}

\subsection{Additional Analysis}

\subsubsection{Employing our framework in pixel space \textit{vs.} latent space}

The proposed \ours is employed in pixel space rather than in latent space produced by autoencoders commonly adopted in previous GVC methods. In Section~\ref{subsec:ablation}, we conduct an ablation study in which residual coding is instead performed in latent space by inserting \ours between the VAE encoder and decoder. Here, we provide the implementation details.

Following~\citep{gnvc-vd}, we adopt the VAE from Wan2.1~\citep{wan} as the autoencoder, which performs $8\times$ spatial downsampling for all frames and $4\times$ temporal downsampling for inter-frames only. As described in Section~\ref{subsec:framework}, $g_\mathrm{a}$ transforms the to-be-encoded $\hat{\mathbf{R}}_t$ into $\hat{\mathbf{y}}_t$ for subsequent entropy coding. The spatial resolution of $\hat{\mathbf{y}}_t$ is reduced by a factor of 16 relative to $\hat{\mathbf{R}}_t$ through an $8 \times 8$ patchification and a convolutional layer with $2\times$ downsampling within $g_\mathrm{a}$. To shift \ours to latent space, we encode two latent frames into $\hat{\mathbf{y}}_t$, corresponding to eight frames as a GOP in pixel space, and remove the patchification to ensure that $\hat{\mathbf{y}}_t$ retains the same dimensions as in pixel-space residual coding. Besides, $16 \times 16$ patchification in the VideoPiT is replaced with $2 \times 2$ patchification.

We provide additional experimental results in Table~\ref{tab:residual coding and residual generation in latent space or pixel space} to evaluate the effectiveness of performing residual coding and residual generation in latent space \textit{vs.} pixel space. A comparison between line (a) and line (b) shows that when five-step residual coding is performed in latent space, compression performance declines, particularly in terms of BD-rate \wrt PSNR. By contrast, comparing line (c) with line (d), the latent-space variant outperforms the pixel-space counterpart on perceptual metrics, \ie, LPIPS, DISTS, and FloLPIPS. This may indicate that under a fixed model capacity, compact residual latent representations are more amenable to generation than pixel-space signals. Comparing line (a) with line (c), both of which operate in latent space, residual generation proves considerably more tractable than residual coding. This might be because compact latent representations are more susceptible to quantization, while quantization errors in latent space may be magnified in pixel space by the VAE decoder. Overall, as evidenced by the comparison between line (e) and line (f), the pixel-space model surpasses the latent-space model under our default setting of two-step coding and three-step generation.

\subsubsection{Distribution of Bitrate Cost across Different Residual Coding Steps}

\begin{figure*}[t]
    \centering
    \includegraphics[width=\textwidth]{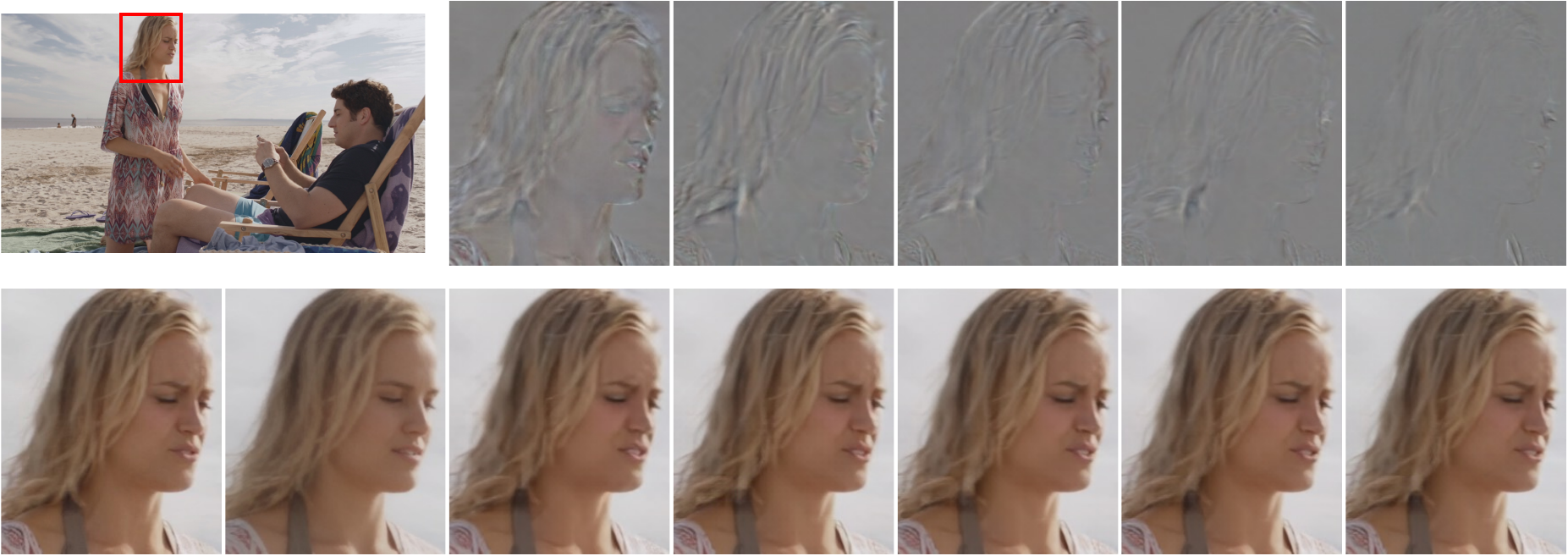}
    \vspace{5pt}
    {\tiny
        \begin{tabularx}{\textwidth}{
          >{\centering\arraybackslash\hsize=1\hsize\linewidth=\hsize}X
          >{\centering\arraybackslash\hsize=1\hsize\linewidth=\hsize}X
          >{\centering\arraybackslash\hsize=1\hsize\linewidth=\hsize}X
          >{\centering\arraybackslash\hsize=1\hsize\linewidth=\hsize}X
          >{\centering\arraybackslash\hsize=1\hsize\linewidth=\hsize}X
          >{\centering\arraybackslash\hsize=1\hsize\linewidth=\hsize}X
          >{\centering\arraybackslash\hsize=1\hsize\linewidth=\hsize}X
        }
        \makecell{Original \\ BPP$\downarrow$ \\ LPIPS$\downarrow$ } & \makecell{Step 0 \\ 0.0188 \\ 0.0725}& \makecell{Step 1 \\ 0.0216 (+0.0028) \\ 0.0453 (-0.0272)} & \makecell{Step 2 \\ 0.0227 (+0.0011) \\ 0.0394 (-0.0059)} & \makecell{Step 3 \\ 0.0230 (+0.0003) \\ 0.0369 (-0.0025)} & \makecell{Step 4 \\ 0.0231 (+0.0001) \\ 0.0354 (-0.0015)} & \makecell{Step 5 \\ 0.0232 (+0.0001) \\ 0.0340 (-0.0014)} \\
        \end{tabularx}
    }
    \vspace{-10pt}
    \caption{Visual comparisons of the reconstructed noisy states during five-step residual coding for the \textit{videoSRC23} sequence from the MCL-JCV dataset. Wherein, residual maps depicted above are derived by subtracting the reconstruction of the preceding step from that of the current step.} 
    \label{fig:visualization on videoSRC23}
\end{figure*}

\begin{table}[t]
\caption{Distribution of bitrate cost across different residual coding steps. Note that the percentage in ``Step 0'' denotes the proportion of the total bitrate consumed by the learned video codec.
}
\centering
\vspace{-7.5pt}
\renewcommand{\arraystretch}{1.05}
\renewcommand{\tabcolsep}{15pt}
\begin{adjustbox}{max width=0.75\linewidth}
    \begin{tabular}{c|c|c|c|c|c}
    \toprule
    Step 0 & Step 1 & Step 2 & Step 3 & Step 4 & Step 5 \\
    \midrule
    82.9\% & 10.1\% & 4.7\% & 1.8\% & 0.3\% & 0.2\% \\ 
    \bottomrule
    \end{tabular}
\end{adjustbox}
\label{tab:distribution of bitrate cost}
\vspace{-5pt}
\end{table}

Table~\ref{tab:distribution of bitrate cost} presents the distribution of bitrate cost across different residual-coding steps. The learned video codec accounts for the largest portion of the total bitrate. During residual coding, the combined bitrate contribution of Step 3, Step 4, and Step 5 is 2.3\% of the total bitrate. Notably, as shown in Table~\ref{tab:different steps for residual coding and residual generation}, taking our default setting (two residual-coding steps and three residual-generation steps) as the anchor, using five residual-coding steps leads to more than a 3\% BD-rate increase under all metrics. This discrepancy may stem from the fact that residual coding in the later steps introduces quantization errors that degrade reconstruction quality, while the residuals generated by VideoPiT are already sufficiently accurate at these denoising steps.

Furthermore, we provide a visual example produced with five-step residual coding to illustrate the distribution of bitrate cost across different steps and the corresponding variation in perceptual quality. As shown in Fig.~\ref{fig:visualization on videoSRC23}, the first two residual-coding steps consume the most bitrate and yield significant perceptual quality improvements. At step 3, step 4, and step 5, perceptual quality continues to improve through residual generation using predicted residual segments.

\subsection{More Visual Examples}

\begin{figure*}[t]
\centering
\includegraphics[width=\textwidth]{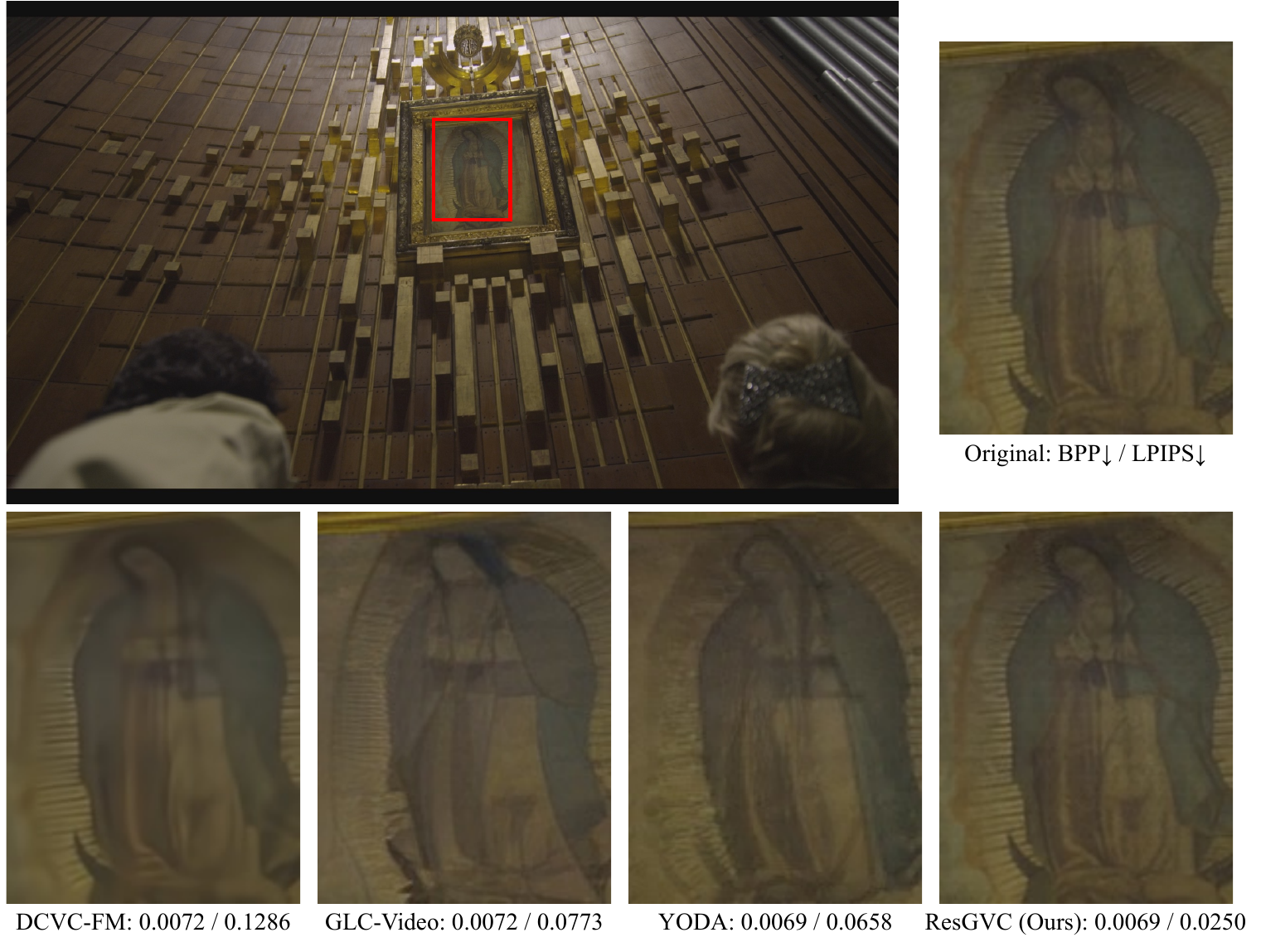}
\vspace{-20pt}
\caption{Visual comparisons with open-source baselines on the \textit{videoSRC03} sequence from the MCL-JCV dataset.}
\vspace{-10pt}
\label{fig:videoSRC03}
\end{figure*}

\begin{figure*}[t]
\centering
\includegraphics[width=\textwidth]{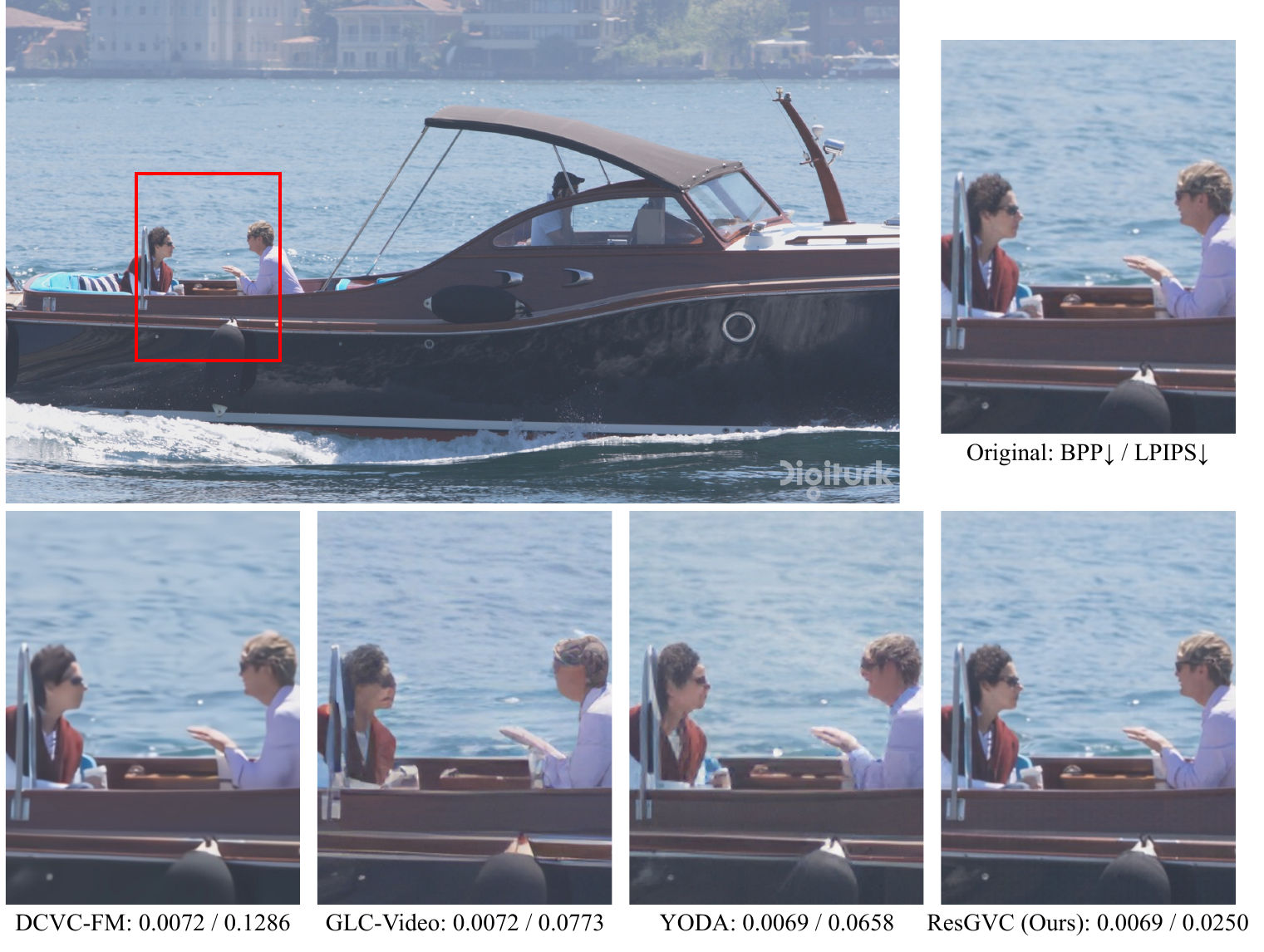}
\vspace{-20pt}
\caption{Visual comparisons with open-source baselines on the \textit{YachtRide} sequence from the UVG dataset.}
\vspace{-15pt}
\label{fig:YachtRide}
\end{figure*}

\begin{figure*}[t]
\centering
\includegraphics[width=\textwidth]{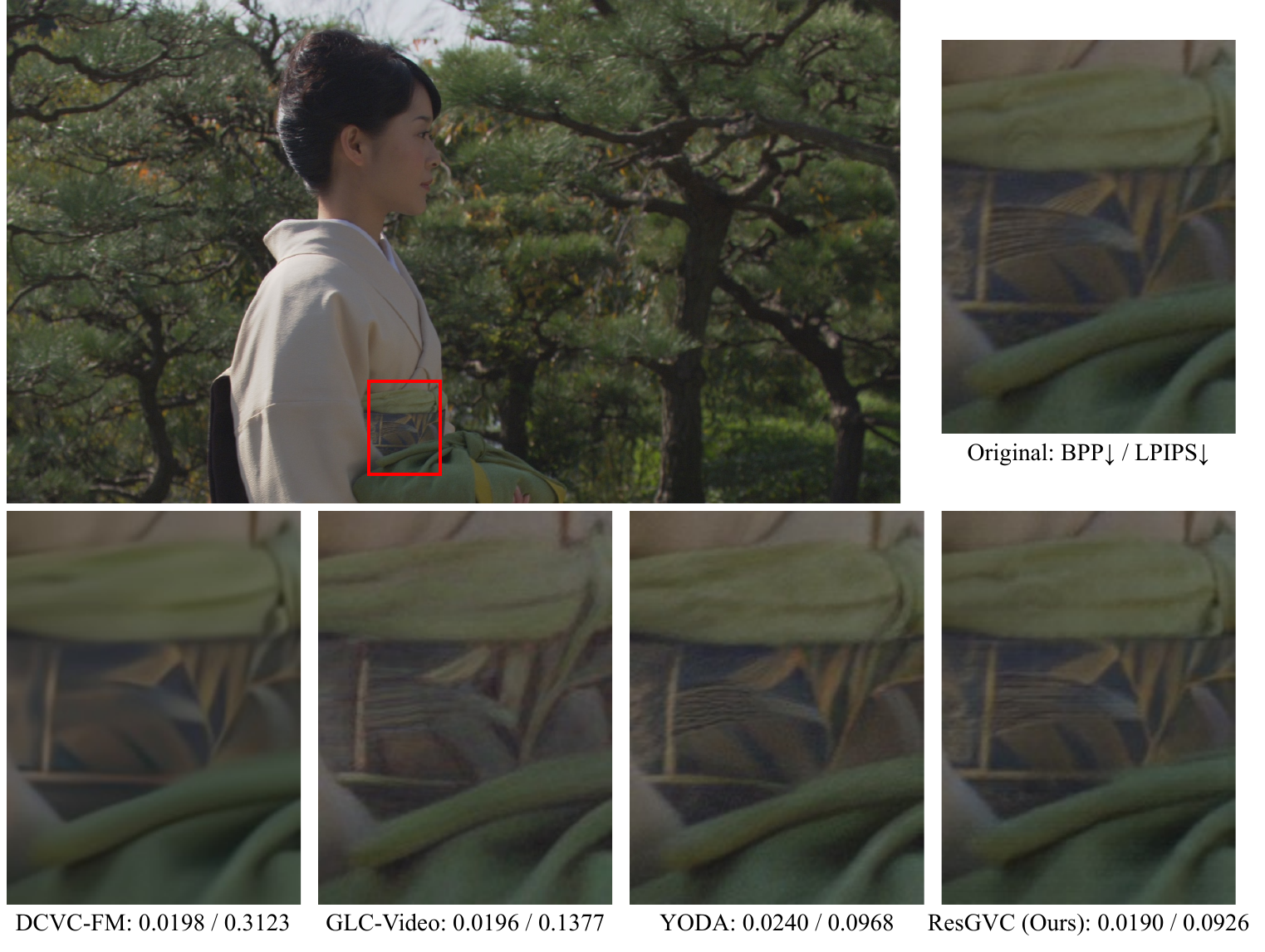}
\vspace{-20pt}
\caption{Visual comparisons with open-source baselines on the \textit{Kimono} sequence from the HEVC B dataset.}
\vspace{-15pt}
\label{fig:Kimono}
\end{figure*}

\begin{figure*}[t]
\centering
\includegraphics[width=\textwidth]{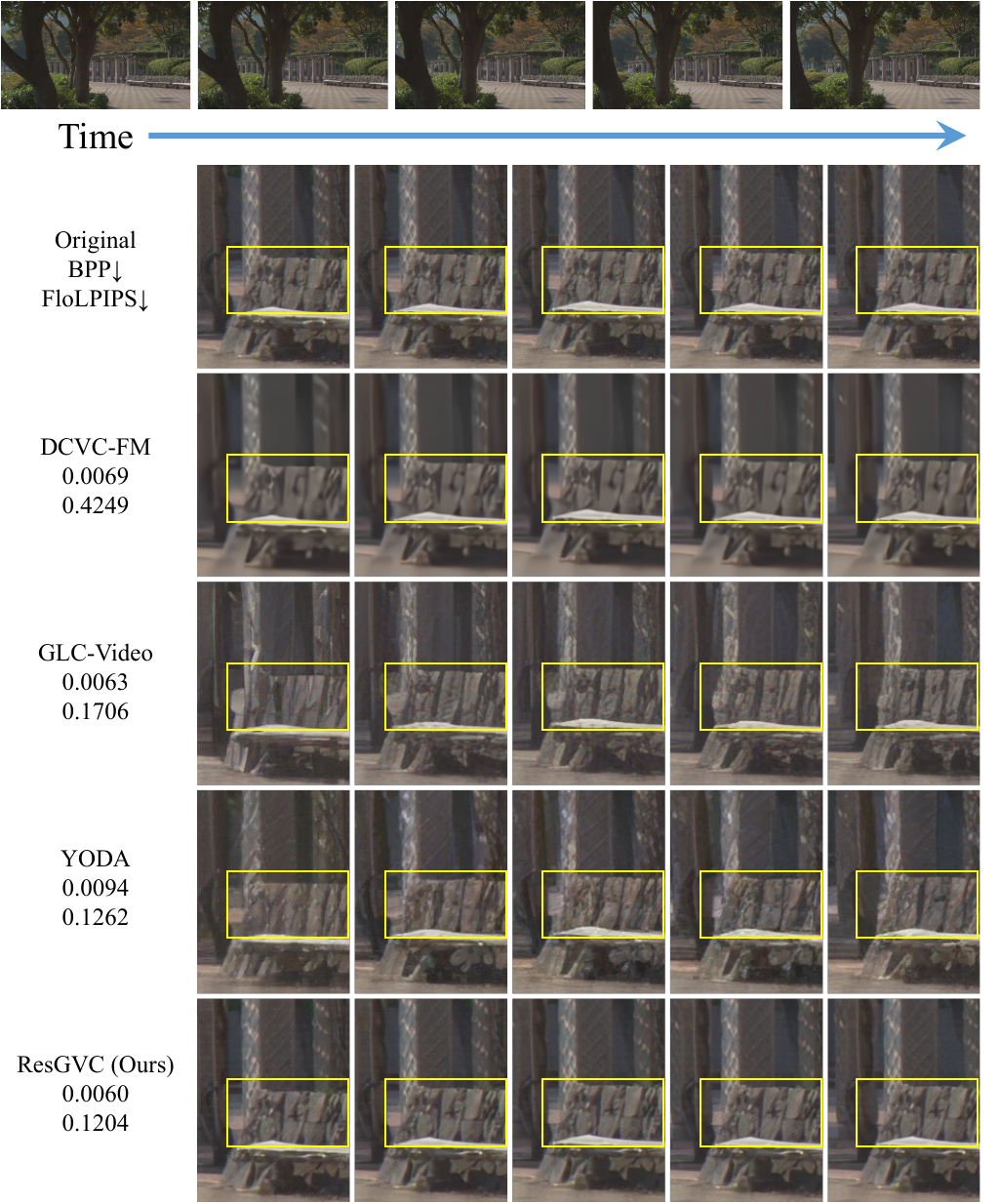}
\vspace{-10pt}
\caption{Visual comparisons across frames on the \textit{ParkScene} sequence from the HEVC B dataset.}
\vspace{-10pt}
\label{fig:ParkScene}
\end{figure*}

We present additional qualitative comparisons on three datasets: HEVC B, MCL-JCV, and UVG. As illustrated in Fig.~\ref{fig:videoSRC03} - Fig.~\ref{fig:Kimono}, our method yields visually sharper and more realistic reconstructions than the competing baselines. Fig.~\ref{fig:ParkScene} presents a comparison of temporal consistency across different methods. DCVC-FM generally produces blurry reconstructions, while GLC-Video and YODA tend to introduce jitter and flickering artifacts, thereby compromising temporal coherence. By contrast, our method delivers temporally stable reconstructions, especially in regions containing dense textures. These examples further underscore that our approach excels not only on quantitative metrics but also in preserving fine textures, realistic details, and perceptual naturalness, thereby delivering superior visual quality.

\end{document}